\documentclass{article}

\PassOptionsToPackage{numbers, sort&compress}{natbib}
\usepackage[final]{neurips_2018}
\usepackage{subcaption}

\usepackage[utf8]{inputenc} %
\usepackage[T1]{fontenc}    
\usepackage{hyperref}       
\usepackage{url}            
\usepackage{booktabs}       
\usepackage{amsfonts}       
\usepackage{nicefrac}       
\usepackage{microtype}      

\usepackage{tabularx}
\usepackage{subcaption}
\usepackage{multirow}
\usepackage{graphicx}

\usepackage{enumitem}

\usepackage{amsthm,amsmath,amssymb}
\usepackage{shortcuts} 
\theoremstyle{plain}
\newtheorem{theorem}{Theorem}
\newtheorem{lemma}{Lemma}
\newtheorem{corollary}[theorem]{Corollary}
\theoremstyle{definition}
\newtheorem{assumption}{Assumption}
\newtheorem{definition}{Definition}

\newtheorem{example}{Example}

\newcommand{\Xs}{{X_{1:n}}}
\newcommand{\Ts}{{T_{1:n}}}
\newcommand{\XTs}{{\Xs,\Ts}}
\newcommand{\Ys}{{Y_{1:n}}}

\renewcommand{\argmin}{\operatornamewithlimits{argmin}}

\newenvironment{talign}
 {\align}
 {\endalign}
\newenvironment{talign*}
 {\csname align*\endcsname}
 {\endalign}

\newcommand{\tendl}{\\}
\newcommand{\ts}{\textstyle}

\newcommand{\logging}{\varphi}

\usepackage[compact]{titlesec}  
\titlespacing\section{0pt}{0.4em plus 0.2em minus 0.2em}{0.4em plus 0.2em minus 0.2em}
\titlespacing\subsection{0pt}{0.4em plus 0.2em minus 0.2em}{0.4em plus 0.2em minus 0.2em}
\titlespacing\subsubsection{0pt}{0.4em plus 0.2em minus 0.2em}{0.4em plus 0.2em minus 0.2em}

\title{Balanced Policy Evaluation and Learning}
\author{
  Nathan Kallus\\
  Cornell University and Cornell Tech\\
  \texttt{kallus@cornell.edu}
}

\begin{document} 

\maketitle
\begin{abstract} 
We present a new approach to 
the problems of evaluating and learning personalized decision policies from
observational data of past contexts, decisions, and outcomes. Only the outcome of the
enacted decision is available and the historical policy is unknown.
These problems arise in personalized medicine 
using electronic health records and in internet advertising.
Existing approaches use inverse propensity weighting (or, doubly robust versions) 
to make historical outcome (or, residual) data look like it were
generated by a new policy being evaluated or learned. But this relies on a plug-in
approach that rejects data points with a decision that disagrees with the new policy,
leading to high variance estimates and ineffective learning. 
We propose a new, balance-based approach that
too makes the data look like the new policy but
does so \emph{directly} by finding weights that
\emph{optimize} for balance between the weighted data and the target policy 
in the given, finite sample, which is equivalent to minimizing
worst-case or posterior conditional mean square error. 
Our policy learner proceeds as a two-level optimization problem over
policies and weights.
We demonstrate that this approach markedly outperforms
existing ones both in evaluation and learning,
which is
unsurprising given the wider support of balance-based weights.
We establish extensive theoretical consistency guarantees 
and regret bounds that support this empirical success.
\end{abstract} 
\section{Introduction}
Using observational data with partially
observed outcomes to develop new and effective 
personalized decision policies
has received increased attention recently 
\citep{dudik2011doubly,li2011unbiased,beygelzimer2009offset,athey2017efficient,
swaminathan2015counterfactual,swaminathan2015self,bottou2013counterfactual,
kallus2016recursive,strehl2010learning,zhou2017residual}.
The aim is to transform electronic health records to 
personalized treatment regimes \citep{bertsimas2017personalized}, transactional records to personalized
pricing strategies \citep{bertsimas2016power},
and click- and ``like''-streams to 
personalized advertising campaigns \citep{bottou2013counterfactual} -- problems of great practical significance.
Many of the existing methods rely on a reduction to weighted classification via
a rejection and importance sampling technique related to inverse propensity weighting
and to doubly robust estimation.
However, inherent in this reduction are several shortcomings that lead
to reduced personalization efficacy:
it involves a na\"ive plug-in estimation of a denominator nuisance parameter
leading either to high variance or scarcely-motivated stopgaps;
it necessarily rejects a significant amount of observations leading to smaller
datasets in effect; and it proceeds in a two-stage approach that is unnatural
to the single learning task.

In this paper, we attempt to ameliorate these by using a new approach that
directly optimizes for the balance that is achieved only on average
or asymptotically by the rejection and importance sampling approach.
We demonstrate that this new approach provides improved performance
and explain why. And, we provide extensive theory to characterize
the behavior of the new methods.
The proofs are given in the supplementary material.

\subsection{Setting, Notation, and Problem Description}

The problem we consider is how to choose the best of $m$ treatments based on an 
observation of 
covariates $x\in\mathcal X\subseteq\R d$ (also known as a context). 
An instance is characterized by the random variables $X\in\mathcal X$ and $Y(1),\dots,Y(m)\in\Rl$, 
where $X$ denotes the covariates and $Y(t)$ for $t\in[m]=\fbraces{1,\dots,m}$ is the 
outcome that would be derived from applying treatment $t$. 
We always assume that \emph{smaller} outcomes
are preferable, \ie, $Y(t)$ corresponds to costs or \emph{negative} rewards.

A policy is a map $\pi:\mathcal X\to\Delta^m$ from observations of covariates 
to a probability vector
in the $m$-simplex $\Delta^m=\fbraces{p\in\R m_+:\sum_{t=1}^mp_t=1}$. Given an observation
of covariates $x$, the policy $\pi$ specifies that treatment $t$ should be applied with 
probability $\pi_t(x)$.
There are two key tasks of interest: policy evaluation and policy learning.
In policy evaluation, we wish to evaluate the performance of a given policy based
on historical data. This is also known as \emph{off}-policy evaluation, highlighting the
fact that
the historical data was not necessarily generated by the policy in question.
In policy learning, we wish to determine a policy that has good performance.

We consider doing both tasks based on data consisting of $n$ 
\emph{passive, historical observations} of covariate, treatment,
and outcome: $S_n=\braces{(X_1,T_1,Y_1),\dots,(X_n,T_n,Y_n)}$, 
where the observed outcome $Y_i=Y_i(T_i)$ corresponds \emph{only} to the treatment 
$T_i$ historically applied. 
We use the notation $X_{1:n}$ to denote the data tuple $(X_1,\dots,X_n)$.
The data is assumed to be iid. That is, the data is generated by drawing from a stationary 
population of instances $(X,T,Y(1),\dots,Y(m))$ and observing a censored form of this draw
given by $(X,T,Y(T))$.%
\footnote{Thus, although the data is iid, the $t$-treated sample $\fbraces{i:T_i=t}$ 
may differ
\emph{systematically} from the $t'$-treated sample $\fbraces{i:T_i=t'}$ for $t\neq t'$, 
\ie, not necessarily just by chance as in a randomized controlled trial (RCT).}
From the (unknown) joint distribution of $(X,T)$ in the population, we define the
(unknown) propensity function $\logging_t(x)=\fPrb{T=t\mid X=x}=\Efb{\delta_{Tt}\mid X=x}$, where $\delta_{st}=\indic{s=t}$ is the Kronecker delta.
And, from the (unknown) joint distribution of $(X,Y(t))$ in the population, we define the
(unknown) mean-outcome function $\mu_t(x)=\Efb{Y(t)\mid X=x}$. 
We use the notation
$\logging(x)=(\logging_1(x),\dots,\logging_m(x))$ and $\mu(x)=(\mu_1(x),\dots,\mu_m(x))$.

Apart from being iid, we also assume the data satisfies unconfoundedness:
\begin{assumption}
\label{unconfoundedness}
For each $t\in[m]$: 
$Y(t)$ is independent of $T$ given $X$,
\ie,
$Y(t)\indep T\mid X$.
\end{assumption}
This assumption is equivalent to there being a \emph{logging policy} 
$\logging$ that generated the data by prescribing
treatment $t$ with probability $\logging_t(X_i)$ to each instance $i$ and recording 
the outcome $Y_i=Y_i(T_i)$.
Therefore, especially in the case where the logging policy $\logging_t$ 
is in fact \emph{known} to the user, 
the problem is often called learning from logged bandit feedback 
\citep{strehl2010learning,swaminathan2015counterfactual}.

In policy evaluation, given a policy $\pi$, we wish to estimate its 
\emph{sample-average policy effect} (SAPE),
$$\ts
\op{SAPE}(\pi)=\frac1n\sum_{i=1}^n\sum_{t=1}^m\pi_t(X_i)\mu_t(X_i),
$$
by an estimator $\hat\tau(\pi)=\hat\tau(\pi;X_{1:n},T_{1:n},Y_{1:n})$ that depends
only on the observed data and the policy $\pi$.
The SAPE quantifies the average outcome that a policy $\pi$ would induce in the
sample and hence measures its risk.
SAPE is
strongly consistent
for
the \emph{population-average policy effect} (PAPE):
$$\ts
\op{PAPE}(\pi)=\Efb{\op{SAPE}(\pi)}=
\Efb{\sum_{t=1}^m\pi_t(X)\mu_t(X)}=\Efb{Y(\tilde T_{\pi(X)})},
$$
where $\tilde T_{\pi(x)}$ is defined as $\pi$'s random draw of treatment when $X=x$,
${\tilde T_{\pi(x)}}\sim \op{Multinomial}(\pi(x))$.
Moreover, if $\pi^*$ is such that $\pi^*_t(x)>0\iff t\in\argmin_{s\in[m]}\mu_s(x)$,
then $\widehat R(\pi)=\op{SAPE}(\pi)-\op{SAPE}(\pi^*)$ is the 
\emph{regret} of $\pi$ \citep{bubeck2012regret}.
The policy evaluation task is closely related to causal effect estimation
\citep{imbens2015causal} where, for $m=2$, one is interested in estimating
the sample and population average treatment effects:
$\op{SATE}=\frac1n\sum_{i=1}^n(\mu_2(X_i)-\mu_1(X_i))$, 
$\op{PATE}=\Efb{\op{SATE}}=\Efb{Y(2)-Y(1)}$.

In policy learning, we wish to find a policy $\hat\pi$ that achieves small outcomes, \ie,
small SAPE and PAPE. The optimal policy $\pi^*$ minimizes both $\op{SAPE}(\pi)$ 
and $\op{PAPE}(\pi)$ 
over all functions $\mathcal X\to\Delta^m$. 

\subsection{Existing Approaches and Related Work}\label{existingsec}

The so-called ``direct'' approach
fits regression
estimates $\hat\mu_t$ of $\mu_t$ on each
dataset $\fbraces{(X_i,Y_i):T_i=t}$, $t\in[m]$. Given these estimates, 
it estimates SAPE in a \emph{plug-in} fashion:
$$\ts
\hat\tau^\text{direct}(\pi)=\frac1n\sum_{i=1}^n\sum_{t=1}^m\pi_t(X_i)\hat\mu_t(X_i).
$$
A policy is learned
either by $\hat\pi^\text{direct}(x)=\argmin_{t\in[m]}\hat\mu_t(x)$
or by minimizing $\hat\tau^\text{direct}(\pi)$ over $\pi\in\Pi$ \citep{qian2011performance}.
However, direct approaches may not generalize as well 
as weighting-based approaches \citep{beygelzimer2009offset}.

Weighting-based approaches seek weights
based on covariate and treatment data $W(\pi)=W(\pi;X_{1:n},T_{1:n})$
that make the outcome data, when reweighted, look
as though it were generated by the policy being evaluated or learned, giving rise to 
estimators that have the form
$$\ts\hat\tau_W=\frac1n\sum_{i=1}^nW_iY_i.$$

\citet{bottou2013counterfactual}, \eg, propose 
to use
inverse propensity weighting (IPW).
Noting 
that \citep{horvitz1952generalization,imbens2000role} 
$\op{SAPE}(\pi)=\Efb{\frac1n\sum_{i=1}^nY_i\times{\pi_{T_i}(X_i)}/{\logging_{T_i}(X_i)}\mid\Xs}$,
one first fits a probabilistic classification model $\hat\logging$ to
$\fbraces{(X_i,T_i):i\in[n]}$
and then estimates SAPE
in an alternate but also \emph{plug-in} fashion:
$$\ts
\hat\tau^\text{IPW}(\pi)=\hat\tau_{W^\text{IPW}(\pi)},\quad W_i^\text{IPW}(\pi)=\pi_{T_i}(X_i)/\hat\logging_{T_i}(X_i)
$$
For a deterministic policy, $\pi_t(x)\in\fbraces{0,1}$, 
this can be interpreted as a rejection and importance sampling approach 
\citep{li2011unbiased,strehl2010learning}:
reject samples where the observed treatment does not match $\pi$'s recommendation
and up-weight those that do by the inverse (estimated) propensity.
For deterministic policies $\pi_t(x)\in\fbraces{0,1}$, we have that
$\pi_{T}(X)=\delta_{T,\tilde T_{\pi(X)}}$ is the complement of 0-1 loss
of $\pi(X)$ in predicting $T$. By scaling and constant shifts, one can therefore
reduce minimizing $\hat\tau^\text{IPW}(\pi)$ over
policies $\pi\in\Pi$ to minimizing a
\emph{weighted classification} loss over \emph{classifiers} $\pi\in\Pi$,
providing a reduction to weighted classification
\citep{beygelzimer2009offset,zhou2017residual}.

Given both $\hat\mu(x)$ and $\hat\logging(x)$ estimates,
\citet{dudik2011doubly} propose a weighting-based approach 
that combines the direct and IPW approaches by 
adapting
the doubly robust (DR) estimator 
\citep{robins1994estimation,robins2000robust,scharfstein1999adjusting,chernozhukov2016double}:
$$\ts\hat\tau^\text{DR}(\pi)=
\frac1n\sum_{i=1}^n\sum_{t=1}^m\pi_t(X_i)\hat\mu_t(X_i)+
\frac1n\sum_{i=1}^n(Y_i-\hat\mu_{T_i}(X_i)){\pi_{T_i}(X_i)}/{\hat\logging_{T_i}(X_i)}.
$$
$\hat\tau^\text{DR}(\pi)$ 
can be understood either as \emph{debiasing} the direct estimator by via
the reweighted residuals $\hat\epsilon_i=Y_i-\hat\mu_{T_i}(X_i)$ or as \emph{denoising}
the IPW estimator by subtracting the conditional mean from $Y_i$. 
As its bias is multiplicative in the biases of the regression and propensity estimates, 
the estimator
is consistent so long as one of the estimates is consistent.
For policy learning, \citep{dudik2011doubly,athey2017efficient} minimize
this estimator via weighted classification. 
\citet{athey2017efficient} provide a tight and favorable 
analysis of the corresponding uniform consistency (and hence regret) of the DR approach
to policy learning.

Based on the fact that $1=\Efb{{\pi_T(X)}/{\logging_{T}(X)}}$,
a normalized IPW (NIPW) estimator is 
given by normalizing the weights so they sum to $n$,
a common practice in causal effect estimation
\citep{lunceford2004stratification,austin2015moving}:
$$\ts
\hat\tau^\text{NIPW}(\pi)=\hat\tau_{W^\text{NIPW}(\pi)},\quad W_i^\text{NIPW}(\pi)=W_i^\text{IPW}(\pi)/\sum_{i'=1}^nW_{i'}^\text{IPW}(\pi).
$$
Any IPW approaches are subject to considerable variance because the plugged-in 
propensities are in the denominator so that small errors 
can have \emph{outsize} effects on the total estimate. Another stopgap
measure is to \emph{clip} the propensities \citep{ionides2008truncated,elliott2008model}
resulting in the clipped IPW (CIPW) estimator:
$$\ts
\hat\tau^\text{$M$-CIPW}(\pi)=\hat\tau_{W^\text{$M$-CIPW}(\pi)},\quad W_i^\text{$M$-CIPW}(\pi)=\pi_{T_i}(X_i)/\max\fbraces{M,\hat\logging_{T_i}(X_i)}.
$$
While effective in reducing variance, the practice remains ad-hoc, loses the unbiasedness
of IPW (with true propensities), and requires the tuning of $M$.
For policy learning, \citet{swaminathan2015counterfactual} propose 
to minimizes over $\pi\in\Pi$ the $M$-CIPW estimator
plus a regularization term of the sample variance of the estimator,
which they term POEM.
The sample variance scales with the level of overlap between 
$\pi$ and $\Ts$, \ie, the prevalence of $\pi_{T_i}(X_i)>0$.
Indeed, when the policy class $\Pi$ is very flexible 
relative to $n$ and if outcomes are 
nonnegative, then the \emph{anti}-logging policy $\pi_{T_i}(X_i)=0$
minimizes any of the above estimates.
POEM avoids learning the anti-logging
policy by regularizing overlap,
reducing variance but limiting novelty of $\pi$.
A refinement, SNPOEM \citep{swaminathan2015self}
uses a normalized \emph{and} clipped IPW (NCIPW) estimator (and regularizes variance):
$$\ts
\hat\tau^\text{$M$-NCIPW}(\pi)=\hat\tau_{W^\text{$M$-NCIPW}(\pi)},\quad W_i^\text{$M$-NCIPW}(\pi)=W_i^\text{$M$-CIPW}(\pi)/\sum_{i'=1}^nW_{i'}^\text{$M$-CIPW}(\pi).
$$
\citet{kallus2018policy} generalize the IPW approach to a continuum of treatments. \citet{kallus2018confounding} suggest a minimax approach to perturbations of the weights to account for confounding factors.
\citet{kallus2016recursive} proposes a recursive partitioning approach to policy learning,
the Personalization Tree (PT) and Personalization Forest (PF),
that dynamically learns both weights and policy, but
still uses within-partition 
IPW with dynamically estimated propensities.

\subsection{A Balance-Based Approach}

\textbf{Shortcomings in existing approaches.}
All of the above weighting-based approaches seek to reweight the historical data so that
they look as though they were generated by the policy being evaluated or learned. Similarly,
the DR approach seeks to make the historical \emph{residuals} look like those that
would be generated under the policy in question so to remove bias from the 
estimated regression model of the direct approach. However, the way these methods
achieve this through various forms and versions of inverse propensity weighting,
has three critical shortcomings:
\begin{enumerate}[label=(\arabic*), align=left, leftmargin=*,labelindent=0in,topsep=-1ex,itemsep=0ex,partopsep=0ex,parsep=0ex]
\item
By taking a simple plug-in approach for a nuisance parameter (propensities) that appears 
in the \emph{denominator}, existing weighting-based methods 
are either subject to \emph{very high} variance or must 
rely on scarcely-motivated stopgap measures such as clipping (see also \citep{kang2007demystifying}).
\item
In the case of deterministic policies (such as an optimal policy), existing methods all 
have weights that are multiples of $\pi_{T_i}(X_i)$, which means that one necessarily
\emph{throws away} every data point $T_i$ that does not agree with the new policy recommendation $\tilde T_{\pi(X_i)}$.
This means that one is essentially only using a much smaller dataset than is available,
leading again to higher variance.%
\footnote{
This problem is \emph{unique} to 
policy evaluation and learning -- in causal effect
estimation, the IPW estimator for 
SATE
has nonzero
weights on \emph{all} of the data points.
For policy learning with  
$m=2$, 
\citet{beygelzimer2009offset,athey2017efficient}  
minimize estimates of the form
$\frac12\fprns{\hat\tau(\pi)-\hat\tau(1(\cdot)-\pi)}$ 
with $\hat\tau(\pi)=\hat\tau^\text{IPW}(\pi)$ or $=\hat\tau^\text{DR}(\pi)$.
This evaluates $\pi$ \emph{relative} to the uniformly random policy and
the resulting total weighted sums over $Y_i$ or $\hat\epsilon_i$
have nonzero weights whether $\pi_{T_i}(X_i)=0$ or not.
While a useful approach
for reduction to weighted classification \citep{beygelzimer2009offset}
or invoking semi-parametric theory 
\citep{athey2017efficient}, 
it only works for $m=2$, has no effect on learning 
as the centering correction is constant in $\pi$,
and, for evaluation, is not an estimator for SAPE.
}
\item
The existing weighting-based methods all
proceed in two stages: first estimate propensities
and then plug these in
to a derived estimator (when the logging policy is unknown).
On the one hand, this raises model specification concerns,
and on the other, is unsatisfactory when the task at hand is not inherently
two-staged -- we wish \emph{only} to evaluate or learn policies, not to learn propensities.
\end{enumerate}
\vspace{\parskip}
\textbf{A new approach.}
We propose a balance-based approach that, like the existing weighting-based
methods, also reweights the historical data to make it look as though they were generated by 
the policy being evaluated or learned and potentially 
denoises outcomes in a doubly robust fashion, 
\emph{but rather} than doing so circuitously via a plug-in approach, we do it \emph{directly}
by finding weights that \emph{optimize} for \emph{balance} between the weighted 
data and the target policy in the given, finite sample.

In particular, we formalize balance as a discrepancy between the reweighted historical 
covariate distribution and that induced by the target policy and prove that it is
directly related to the worst-case \emph{conditional mean square error} (CMSE) of 
\emph{any} weighting-based estimator. Given a policy $\pi$, we then propose to choose 
(policy-dependent) weights $W^*(\pi)$ that \emph{optimize} the worst-case CMSE
and therefore achieve excellent balance while controlling for variance. 
For evaluation, we use these optimal weights to evaluate the performance of $\pi$ by
the estimator $\hat\tau_{W^*(\pi)}$ as well as a doubly robust version.
For learning, we propose a \emph{bilevel} optimization problem: minimize over $\pi\in\Pi$,
the estimated risk $\hat\tau_{W^*(\pi)}$ (or a doubly robust version thereof 
and potentially plus a regularization term),
given by the weights $W^*(\pi)$ that minimize the estimation error. 
Our empirical results show the stark benefit of this approach while
our main theoretical results (Thm.~\ref{uniformconsistencythm}, Cor.~\ref{uniformconsistencycor1}) establish vanishing regret bounds.

\section{Balanced Evaluation}

\subsection{CMSE and Worst-Case CMSE}

We begin by presenting the approach in the context of evaluation.
Given a policy $\pi$,
consider \emph{any} weights $W=W(\pi;\XTs)$ 
that are based on the covariate and treatment data.
Given these weights we can consider both a simple weighted estimator 
as well as a
$W$-weighted doubly robust estimator given a regression estimate $\hat\mu$:
$$\ts
\hat\tau_W=\frac1n\sum_{i=1}^nW_iY_i,\quad
\hat\tau_{W,\hat\mu}=
\frac1n\sum_{i=1}^n\sum_{t=1}^m\pi_t(X_i)\hat\mu_t(X_i)+
\frac1n\sum_{i=1}^nW_i(Y_i-\hat\mu_{T_i}(X_i)).
$$

We can measure the risk of either such
estimator as the conditional mean square error (CMSE), conditioned on all of 
the data upon which the chosen weights depend:
$$\ts
\op{CMSE}(\hat\tau,\pi)=\Efb{(\hat\tau-\op{SAPE}(\pi))^2\mid X_{1:n},T_{1:n}}.
$$
Minimal CMSE is the target of choosing weights for weighting-based policy evaluation.
Basic manipulations 
under the unconfoundedness assumption
decompose the CMSE of any weighting-based policy evaluation estimator
into its conditional bias and variance:
\begin{theorem}\label{cmsethm}
Let $\epsilon_{i}=Y_i-\mu_{T_i}(X_i)$ and $\Sigma=\op{diag}(\Efb{\epsilon_{1}^2\mid X_1,T_1},\dots,\Efb{\epsilon_{n}^2\mid X_n,T_n})$. 
Define
$$\ts
B_t(W,\pi_t;f_t)=
\frac1n\sum_{i=1}^n(W_i\delta_{T_it}-\pi_t(X_i))f_t(X_i)\quad\text{and}\quad
B(W,\pi;f)=\sum_{t=1}^m B_t(W,\pi_t;f_t)
$$
\makebox[13em][l]{Then we have that:}\hfill$\hat\tau_{W}-\op{SAPE}(\pi)=
B(W,\pi;\mu)
+\frac1n\sum_{i=1}^n W_i\epsilon_{i}.$\\
\makebox[13em][l]{Moreover, under Asn.~\ref{unconfoundedness}:}\hfill%
$\op{CMSE}(\hat\tau_{W},\pi)=
B^2(W,\pi;\mu)
+\frac1{n^2}W^T\Sigma W.$
\end{theorem}
\begin{corollary}\label{drcor}
Let $\hat\mu$ be given such that $\hat\mu\indep\Ys\mid\XTs$ 
(\eg, trained on a split sample).
\\\makebox[13em][l]{Then we have that:}\hfill$\hat\tau_{W,\hat\mu}-\op{SAPE}(\pi)=
B(W,\pi;\mu-\hat\mu)
+\frac1n\sum_{i=1}^n W_i\epsilon_{i}.$\\
\makebox[13em][l]{Moreover, under Asn.~\ref{unconfoundedness}:}\hfill%
$\op{CMSE}(\hat\tau_{W,\hat\mu},\pi)=
B^2(W,\pi;\mu-\hat\mu)
+\frac1{n^2}W^T\Sigma W.$
\end{corollary}
In Thm.~\ref{cmsethm} and Cor.~\ref{drcor}, 
$B(W,\pi;\mu)$ and $B(W,\pi;\mu-\hat\mu)$ are precisely the conditional \emph{bias} in evaluating $\pi$ 
for $\hat\tau_W$ and $\hat\tau_{W,\hat\mu}$, respectively,
and $\frac1{n^2}W^T\Sigma W$ the conditional \emph{variance} for both.
In particular, $B_t(W,\pi_t;\mu_t)$ or $B_t(W,\pi_t;\mu_t-\hat\mu_t)$
is the conditional {bias} in evaluating the effect
on the instances where $\pi$ assigns $t$. Note that for any function $f_t$, 
$B_t(W,\pi_t;f_t)$ corresponds to the \emph{discrepancy} between the $f_t(X)$-moments
of the measure $\nu_{t,\pi}(A)=\frac1n\sum_{i=1}^n\pi_t(X_i)\indic{X_i\in A}$ 
on $\mathcal X$ and the measure 
$\nu_{t,W}(A)=\frac1n\sum_{i=1}^nW_i\delta_{T_it}\indic{X_i\in A}$. 
The sum $B(W,\pi;f)$ corresponds to the sum of moment
discrepancies over the components of $f=(f_1,\dots,f_m)$ between these measures.
The moment discrepancy of interest is that of $f=\mu$ or $f=\mu-\hat\mu$, but 
neither of these are known.

\begin{figure}[t!]\centering%
\caption{The setting in Ex.~\ref{ex1}}\label{figex1}
\begin{subfigure}[b]{0.25\textwidth}%
\includegraphics[width=\textwidth]{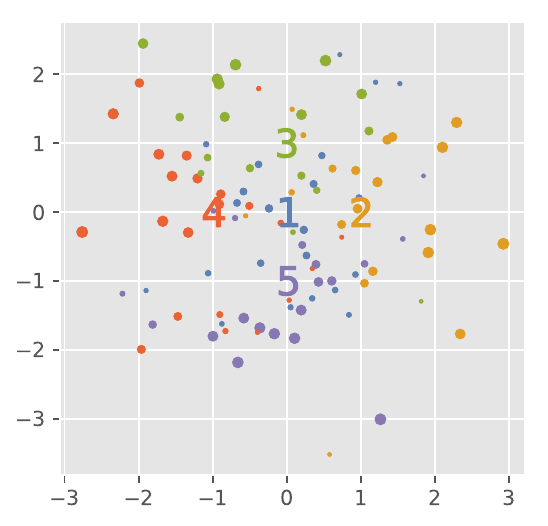}%
\caption{$\XTs$}\label{figex1xplot}%
\end{subfigure}%
\hspace{0.0625\textwidth}\begin{subfigure}[b]{0.25\textwidth}%
\includegraphics[width=\textwidth]{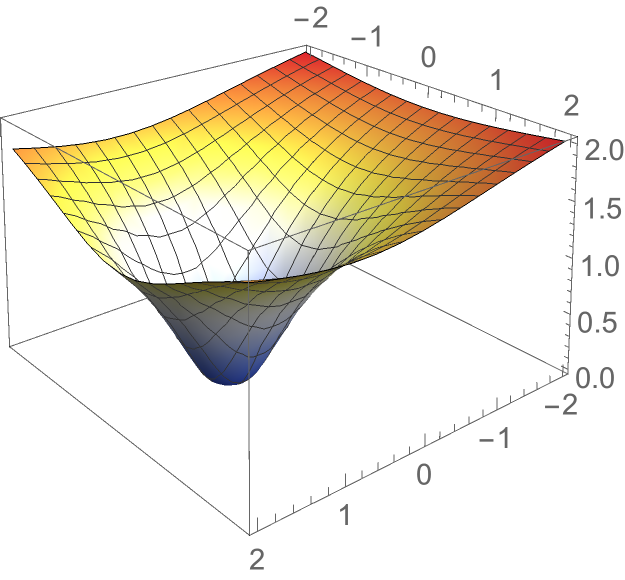}%
\caption{$\mu_1(x)$}\label{figex13dplot}%
\end{subfigure}%
\hspace{0.0625\textwidth}\begin{subfigure}[b]{0.25\textwidth}%
\includegraphics[width=\textwidth]{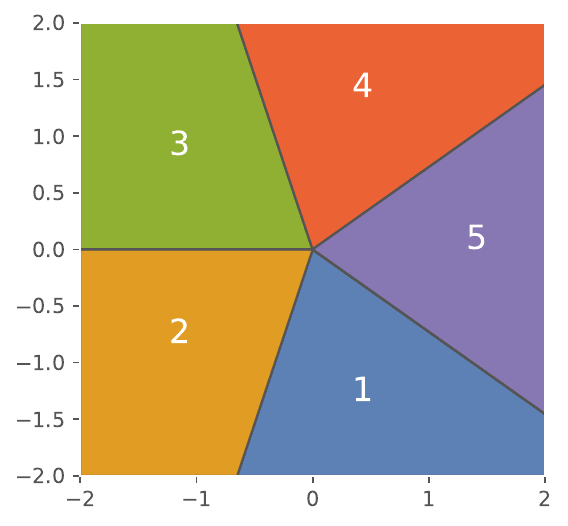}%
\caption{$\pi^*(x)$}\label{figex1bayes}%
\end{subfigure}%
\vspace{-1.25em}
\end{figure}%
\newcommand{\directM}{-0.114}
\newcommand{\directS}{0.308}
\newcommand{\directE}{0.328}
\newcommand{\IPWregtruAM}{-0.005}
\newcommand{\IPWregtruAS}{2.209}
\newcommand{\IPWregtruAE}{2.209}
\newcommand{\IPWregestAM}{-0.491}
\newcommand{\IPWregestAS}{0.310}
\newcommand{\IPWregestAE}{0.581}
\newcommand{\IPWregtruBM}{-0.402}
\newcommand{\IPWregtruBS}{0.329}
\newcommand{\IPWregtruBE}{0.520}
\newcommand{\IPWregestBM}{-0.514}
\newcommand{\IPWregestBS}{0.242}
\newcommand{\IPWregestBE}{0.568}
\newcommand{\IPWnormtruAM}{-0.181}
\newcommand{\IPWnormtruAS}{0.487}
\newcommand{\IPWnormtruAE}{0.519}
\newcommand{\IPWnormestAM}{-0.250}
\newcommand{\IPWnormestAS}{0.415}
\newcommand{\IPWnormestAE}{0.485}
\newcommand{\IPWnormtruBM}{-0.211}
\newcommand{\IPWnormtruBS}{0.405}
\newcommand{\IPWnormtruBE}{0.457}
\newcommand{\IPWnormestBM}{-0.251}
\newcommand{\IPWnormestBS}{0.390}
\newcommand{\IPWnormestBE}{0.463}
\newcommand{\DRregtruAM}{0.435}
\newcommand{\DRregtruAS}{4.174}
\newcommand{\DRregtruAE}{4.196}
\newcommand{\DRregestAM}{0.259}
\newcommand{\DRregestAS}{0.451}
\newcommand{\DRregestAE}{0.520}
\newcommand{\DRregtruBM}{0.267}
\newcommand{\DRregtruBS}{0.432}
\newcommand{\DRregtruBE}{0.508}
\newcommand{\DRregestBM}{0.230}
\newcommand{\DRregestBS}{0.361}
\newcommand{\DRregestBE}{0.428}
\newcommand{\DRnormtruAM}{0.408}
\newcommand{\DRnormtruAS}{0.634}
\newcommand{\DRnormtruAE}{0.754}
\newcommand{\DRnormestAM}{0.471}
\newcommand{\DRnormestAS}{0.550}
\newcommand{\DRnormestAE}{0.724}
\newcommand{\DRnormtruBM}{0.428}
\newcommand{\DRnormtruBS}{0.544}
\newcommand{\DRnormtruBE}{0.692}
\newcommand{\DRnormestBM}{0.467}
\newcommand{\DRnormestBS}{0.511}
\newcommand{\DRnormestBE}{0.692}
\newcommand{\optIPWAM}{0.227}
\newcommand{\optIPWAS}{0.163}
\newcommand{\optIPWAE}{0.280}
\newcommand{\optDRAM}{-0.006}
\newcommand{\optDRAS}{0.251}
\newcommand{\optDRAE}{0.251}
\newcommand{\IPWsupportM}{13.6}
\newcommand{\IPWsupportS}{2.9}
\newcommand{\optsupportM}{90.7}
\newcommand{\optsupportS}{3.2}

\begin{table}[t!]\footnotesize%
\caption{Policy evaluation performance in Ex.~\ref{ex1}}\label{tableex1}%
\centering%
\begin{tabular}{lp{0pt}crrp{0pt}crrp{0pt}r}\toprule
\multicolumn{1}{c}{\multirow{2}{*}{{\centering Weights $W$}}}
&& \multicolumn{3}{c}{Vanilla $\hat\tau_W$} && \multicolumn{3}{c}{Doubly robust $\hat\tau_{W,\hat\mu}$}
&& \multicolumn{1}{c}{\multirow{2}{*}{
$\fmagd W_0$%
}}\\[-.3em]\cmidrule{3-5}\cmidrule{7-9}
&& \multicolumn{1}{c}{RMSE}& \multicolumn{1}{c}{Bias} & \multicolumn{1}{c}{SD} 
&& \multicolumn{1}{c}{RMSE}& \multicolumn{1}{c}{Bias} & \multicolumn{1}{c}{SD}  
\\[-.2em]\midrule
IPW, $\logging$
&& $\IPWregtruAE$ & $\IPWregtruAM$ & $\IPWregtruAS$
&& $\DRregtruAE$  & $\DRregtruAM$  & $\DRregtruAS$
&& $\IPWsupportM\pm\IPWsupportS$
\\
IPW, $\hat\logging$
&& $\IPWregestBE$ & $\IPWregestBM$ & $\IPWregestBS$
&& $\DRregestBE$  & $\DRregestBM$  & $\DRregestBS$
&& $\IPWsupportM\pm\IPWsupportS$
\\
$.05$-CIPW, $\logging$
&& $\IPWregestAE$ & $\IPWregestAM$ & $\IPWregestAS$
&& $\DRregestAE$  & $\DRregestAM$  & $\DRregestAS$
&& $\IPWsupportM\pm\IPWsupportS$
\\
$.05$-CIPW, $\hat\logging$
&& $\IPWregestBE$ & $\IPWregestBM$ & $\IPWregestBS$
&& $\DRregestBE$  & $\DRregestBM$  & $\DRregestBS$
&& $\IPWsupportM\pm\IPWsupportS$
\\
NIPW, $\logging$
&& $\IPWnormtruAE$ & $\IPWnormtruAM$ & $\IPWnormtruAS$
&& $\DRnormtruAE$  & $\DRnormtruAM$  & $\DRnormtruAS$
&& $\IPWsupportM\pm\IPWsupportS$
\\
NIPW, $\hat\logging$
&& $\IPWnormestBE$ & $\IPWnormestBM$ & $\IPWnormestBS$
&& $\DRnormestBE$  & $\DRnormestBM$  & $\DRnormestBS$
&& $\IPWsupportM\pm\IPWsupportS$
\\
$.05$-NCIPW, $\logging$
&& $\IPWnormestAE$ & $\IPWnormestAM$ & $\IPWnormestAS$
&& $\DRnormestAE$  & $\DRnormestAM$  & $\DRnormestAS$
&& $\IPWsupportM\pm\IPWsupportS$
\\
$.05$-NCIPW, $\hat\logging$
&& $\IPWnormestBE$ & $\IPWnormestBM$ & $\IPWnormestBS$
&& $\DRnormestBE$  & $\DRnormestBM$  & $\DRnormestBS$
&& $\IPWsupportM\pm\IPWsupportS$
\\[-.2em]
\cmidrule{1-11}
Balanced eval
&& $\mathbf{\optIPWAE}$ & $\optIPWAM$ & $\optIPWAS$
&& $\mathbf{\optDRAE}$  & $\optDRAM$  & $\optDRAS$
&& $\optsupportM\pm\optsupportS$
\\\bottomrule
\end{tabular}
\end{table}

Balanced policy evaluation 
seeks weights $W$ to minimize a combination of
\emph{imbalance}, given by the worst-case value of $B(W,\pi;f)$ over functions $f$,
and \emph{variance}, given by the norm of weights $W^T\Lambda W$ 
for a specified positive semidefinite (PSD)
matrix $\Lambda$.
This follows a general approach introduced by \citep{kallus2016generalized,kallus2018optimal} of finding optimal balancing weights that optimize a given CMSE objective directly rather than via a plug-in approach.
Any choice of $\fmagd\cdot$ gives rise to a \emph{worst-case CMSE} objective for policy evaluation:
$$\ts
\mathfrak E^2(W,\pi;\fmagd\cdot,\Lambda)
=\sup_{\fmagd{f}\leq1}
B^2(W,\pi;f)
+\frac1{n^2}W^T\Lambda W
.
$$
Here, we focus on $\fmagd\cdot$ given by the 
\emph{direct product of reproducing kernel Hilbert spaces (RKHS)}:
$$\ts\fmagd f_{p,\mathcal K_{1:m},\gamma_{1:m}}=\fprns{\sum_{t=1}^m\fmagd{f_t}_{\mathcal K_t}^p/\gamma_t^p}^{1/p},$$
where $\fmagd\cdot_{\mathcal K_t}$ is the norm of the RKHS given
by the PSD kernel $\mathcal K_t(\cdot,\cdot):\mathcal X^2\to\Rl$,
\ie, the unique completion of $\op{span}(\mathcal K_t(x,\cdot):x\in\mathcal X)$
endowed 
with $\ip{\mathcal K_t(x,\cdot)}{\mathcal K_t(x',\cdot)}=\mathcal K_t(x,x')$
\citep[see][]{scholkopf2001learning}. 
We say $\fmagd f_{\mathcal K_t}=\infty$ if $f$ is not in the RKHS.
One example of a kernel is the Mahalanobis RBF kernel:
$
\mathcal K_s(x,x')=\exp(-(x-x')^T\hat S^{-1}(x-x')/s^2)
$ where $\hat S$ is the sample covariance of $X_{1:n}$ and $s$ is a
parameter.
For such an RKHS product norm, we can decompose the worst-case objective into the discrepancies in each treatment as well as characterize it as a posterior (rather than worst-case) risk.
\begin{lemma}\label{boundlemma}
Let $\mathfrak B^2_t(W,\pi_t;\fmagd\cdot_{\mathcal K_t})=\sum_{i,j=1}^n
(W_i\delta_{T_it}-\pi_t(X_i))(W_j\delta_{T_jt}-\pi_t(X_j))\mathcal K_t(X_i,X_j)$ and
$1/p+1/q=1$. Then
$$\ts
\mathfrak E^2(W,\pi;\fmagd \cdot_{p,\mathcal K_{1:m},\gamma_{1:m}},\Lambda)
=
\fprns{\sum_{t=1}^m\gamma_t^q\mathfrak B_t^q(W,\pi_t;\fmagd\cdot_{\mathcal K_t})}^{2/q}
+\frac1{n^2}W^T\Lambda W.$$
Moreover, if $p=2$ and $\mu_t$ has a 
Gaussian process prior \citep{williams2006gaussian}
with mean $f_t$ and covariance $\gamma_t\mathcal K_t$ then
$$\op{CMSE}(\hat\tau_{W,f},\pi)=\mathfrak E^2(W,\pi;\fmagd \cdot_{p,\mathcal K_{1:m},\gamma_{1:m}},\Sigma),$$
where the CMSE marginalizes over $\mu$. 
This gives the CMSE of $\hat\tau_{W}$ for $f$ constant or $\hat\tau_{W,\hat\mu}$ for $f=\hat\mu$. 
\end{lemma}
The second statement in Lemma~\ref{boundlemma} suggests that, in practice,
model selection
of $\gamma_{1:m}$, $\Lambda$, 
and kernel hyperparameters such as $s$ or even $\hat S$,
can done by the marginal likelihood
method
\citep[see][Ch.~5]{williams2006gaussian}.

\subsection{Evaluation Using Optimal Balancing Weights}

Our policy evaluation estimates
are given by either the estimator 
$\hat\tau_{W^*(\pi;\fmagd\cdot,\Lambda)}$ 
or $\hat\tau_{W^*(\pi;\fmagd\cdot,\Lambda),\hat\mu}$ where 
$W^*(\pi)=W^*(\pi;\fmagd\cdot,\Lambda)$ is the \emph{minimizer} of 
$\mathfrak E^2(W,\pi;\fmagd\cdot,\Lambda)$ over the space of
all weights $W$ that sum to $n$,
$\mathcal W=\fbraces{W\in\R n_+:\sum_{i=1}^nW_i=n}=n\Delta^n$. Specifically,
$$\ts
W^*(\pi;\fmagd\cdot,\Lambda)\in\argmin_{W\in\mathcal W}\mathfrak E^2(W,\pi;\fmagd\cdot,\Lambda).
$$
When $\fmagd \cdot=\fmagd \cdot_{p,\mathcal K_{1:m},\gamma_{1:m}}$,
this problem is a quadratic program for $p=2$ and a
second-order cone program for $p=1,\infty$. Both are 
efficiently solvable \citep{boyd2004convex}. In practice, we solve these using Gurobi 7.0.

In Lemma~\ref{boundlemma}, $\mathfrak B_t(W,\pi_t;\fmagd\cdot_{\mathcal K_t})$
measures the \emph{imbalance} between
$\nu_{t,\pi}$ and $\nu_{t,W}$ as the \emph{worst-case} discrepancy in means over
functions in the unit ball of an RKHS. In fact, as a distributional
distance metric, it is the maximum mean discrepancy (MMD)
used, for example, for testing whether two samples come from the same distribution
\citep{gretton2006kernel}. Thus, minimizing 
$\mathfrak E^2(W,\pi;\fmagd \cdot_{p,\mathcal K_{1:m},\gamma_{1:m}},\Lambda)$
is simply 
seeking the weights $W$ that balance $\nu_{t,\pi}$ and $\nu_{t,W}$
subject to variance regularization in $W$.

\begin{example}\label{ex1}%
We demonstrate balanced evaluation
with a mixture of $m=5$ Gaussians:
$X\mid T\sim\mathcal N\fprns{\overline X_T,I_{2\times 2}}$,
$\overline X_1=(0,0)$,
$\overline X_t=(\op{Re},\op{Im})(e^{i2\pi (t-2)/(m-1)})$ for $t=2,\dots,m$,
and $T\sim\op{Multinomial}(1/5,\dots,1/5)$.
Fix a draw of $\XTs$ with $n=100$ shown in Fig.~\ref{figex1xplot}
(numpy seed 0). Color denotes $T_i$ and size denotes $\logging_{T_i}(X_i)$.
The centers $\overline X_t$ are marked by a colored number.
Next, we let $\mu_t(x)=\exp\fprns{1-1/\fmagd{x-\chi_t}_2}$
where $\chi_t=(\op{Re},\op{Im})(e^{-i2\pi t/m}/\sqrt{2})$ for $t\in[m]$, 
$\epsilon_i\sim\mathcal N(0,\sigma)$, and $\sigma=1$.
Fig.~\ref{figex13dplot} plots $\mu_1(x)$. Fig.~\ref{figex1bayes}
shows the corresponding optimal policy $\pi^*$.

Next we consider evaluating $\pi^*$.
Fixing $\Xs$ as in Fig.~\ref{figex1xplot}, we have $\op{SAPE}(\pi^*)=0.852$.
With $\Xs$ fixed, we draw 1000 replications of $\Ts,\Ys$ from their conditional distribution.
For each replication, we fit $\hat\logging$ by estimating the (well-specified) Gaussian
mixture by maximum likelihood
and fit $\hat\mu$ using $m$ separate gradient-boosted
tree models (sklearn defaults). We consider evaluating $\pi^*$ either using
the vanilla estimator $\hat\tau_W$ or the doubly robust estimator $\hat\tau_{W,\hat\mu}$
for $W$ either chosen in the 4 different standard ways laid out in Sec.~\ref{existingsec},
using either the true $\logging$ or the estimated $\hat\logging$, or chosen by the
balanced evaluation approach using untuned
parameters (rather than fit by marginal likelihood)
using the standard ($s=1$) Mahalanobis 
RBF kernel for $\mathcal K_t$,
$\fmagd f^2=\sum_{t=1}^m\fmagd{f_t}_{\mathcal K_t}^2$, and $\Lambda=I$. 
(Note that this \emph{misspecifies} the outcome model,
$\fmagd{\mu_t}_{\mathcal K_t}=\infty$.)
We tabulate the results in Tab.~\ref{tableex1}.

We note a few observations on the standard approaches:
vanilla IPW with true $\logging$ has zero bias
but large SD (standard deviation) and hence RMSE (root mean square error); 
a DR approach improves on a
vanilla IPW with $\hat\logging$
by reducing bias;
clipping and normalizing IPW reduces SD.
The balanced evaluation approach achieves the best RMSE by a clear
margin, with the vanilla estimator beating all standard vanilla \emph{and} DR
estimators and the DR estimator providing a further improvement by
nearly eliminating bias (but increasing SD). The marked success of the balanced
approach is \emph{unsurprising} when considering the support 
$\fmagd W_0=\sum_{i=1}^n\indic{W_i>0}$ of the weights.
All standard approaches use weights that are multiples of $\pi_{T_i}(X_i)$,
limiting support to the overlap between $\pi$ and $\Ts$, which hovers around $10$--$16$
over replications.
The balanced approach uses weights that have significantly wider
support, around $88$--$94$.
In light of this, the success of the balanced approach is expected.
\end{example}

\subsection{Consistent Evaluation}

Next we consider the question of consistent evaluation: under what conditions
can we guarantee that $\hat\tau_{W^*(\pi)}-\op{SAPE}(\pi)$ and 
$\hat\tau_{W^*(\pi),\hat\mu}-\op{SAPE}(\pi)$ converge to zero and at what rates. 

One key requirement for consistent evaluation is a weak form of overlap between
the historical data and the target policy to be evaluated using this data:
\begin{assumption}[Weak overlap]\label{weakoverlap}\hspace{-.25em}
$\fPrb{\logging_t(X)>0\vee\pi_t(X)=0}=1\,\forall t\in[m]$,
$\Efb{{{\pi^2_T(X)}/{\logging^2_T(X)}}}<\infty$.
\end{assumption}
This ensures that if $\pi$ can assign treatment $t$ to $X$ then the data will have some examples of units with similar covariates being given treatment $t$; otherwise, we can never say what the outcome might look like.
Another key requirement is specification. If the mean-outcome function is well-specified
in that it is in the RKHS product used to compute $W^*(\pi)$ then 
convergence at rate $1/\sqrt{n}$ is guaranteed. 
Otherwise, for a doubly robust estimator, if the regression estimate is well-specified
then consistency is still guaranteed.
In lieu of specification, consistency is also guaranteed
if the RKHS product consists of $C_0$-universal kernels, 
defined below, such as the RBF kernel \citep{sriperumbudur2010universality}.
\begin{definition}
A PSD kernel $\mathcal K$ on a Hausdorff $\mathcal X$ (\eg, $\R d$) is $C_0$-\emph{universal} if, for any continuous function $g:\mathcal X\to\Rl$  with compact support (\ie, for some $C$ compact, $\{x:g(x)\neq 0\}\subseteq C$)
and $\eta>0$, there exists $m,\alpha_1,x_1,\dots,\alpha_m,x_m$ such that $\sup_{x\in\mathcal X}\fabs{\sum_{j=1}^m\alpha_i\mathcal K(x_j,x)-g(x)}\leq\eta$.
\end{definition}
\begin{figure}[t!]\centering\scriptsize%
\caption{Policy learning results in Ex.~\ref{ex2}; numbers denote regret}
\label{figex2}
\parbox{0.28\textwidth}{
\raisebox{-4.95em}{\includegraphics[width=0.28\textwidth]{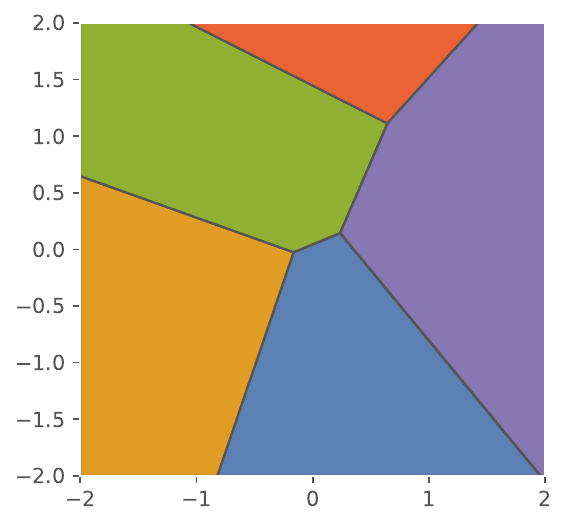}}%
\\\phantom{.}~~~~~~~~Balanced policy learner\hfill.06~~~}%
\bgroup%
\def\arraystretch{0.8}%
\begin{tabular}{cccc}%
\includegraphics[width=0.15\textwidth]{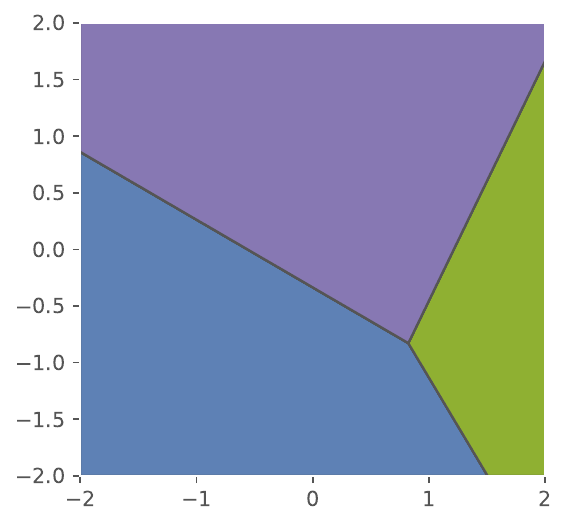}%
&%
\includegraphics[width=0.15\textwidth]{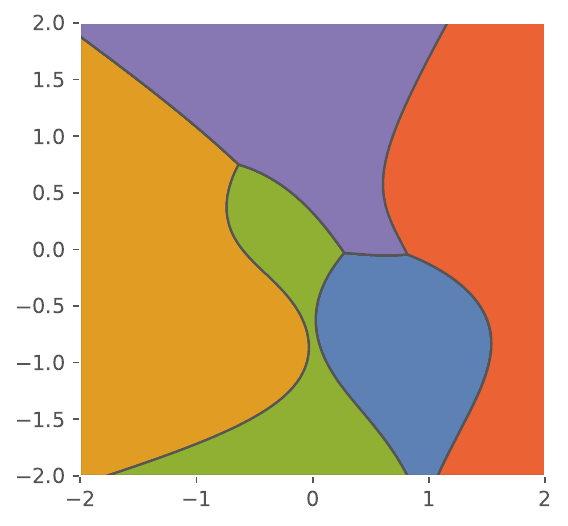}%
&%
\includegraphics[width=0.15\textwidth]{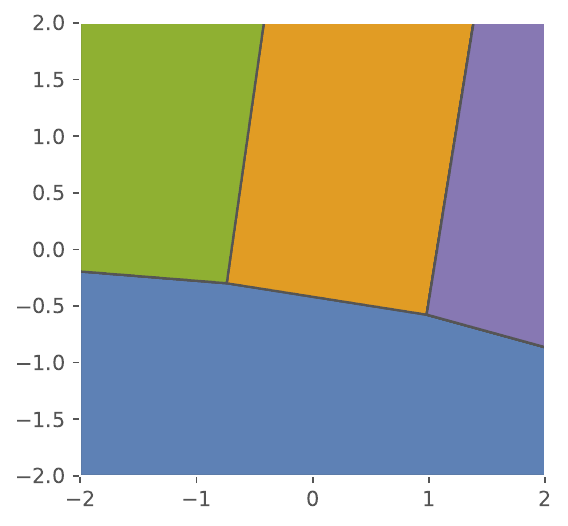}%
&%
\includegraphics[width=0.15\textwidth]{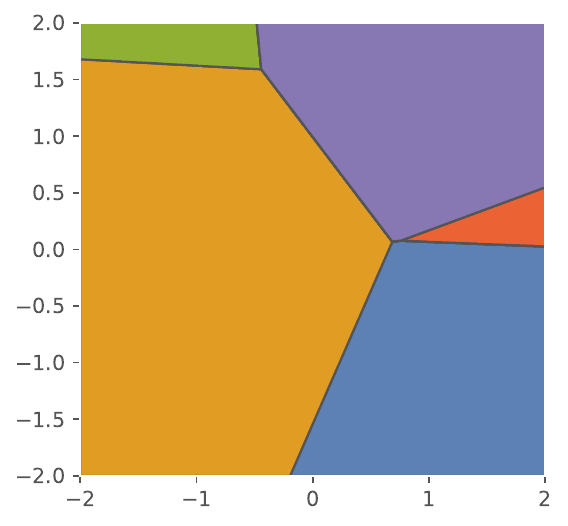}%
\\%
~~~~IPW\hfill.50~&~~~~Gauss Proc\hfill0.29~&~~~~IPW-SVM\hfill0.34~&~~~~SNPOEM\hfill0.28~%
\\%
\includegraphics[width=0.15\textwidth]{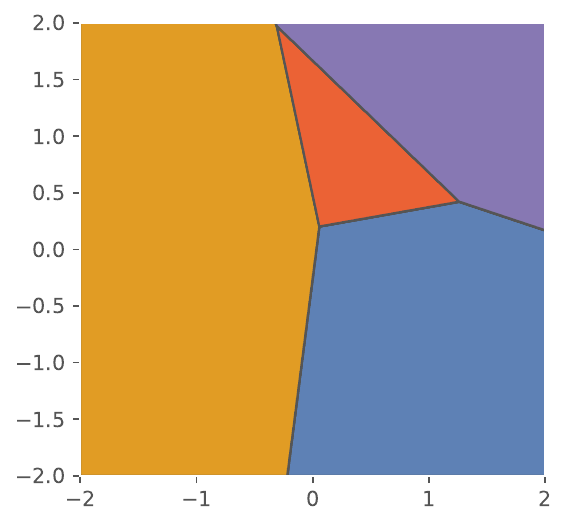}%
&%
\includegraphics[width=0.15\textwidth]{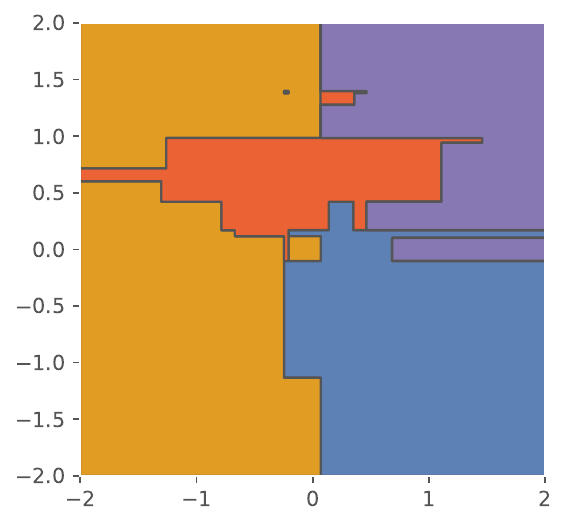}%
&%
\includegraphics[width=0.15\textwidth]{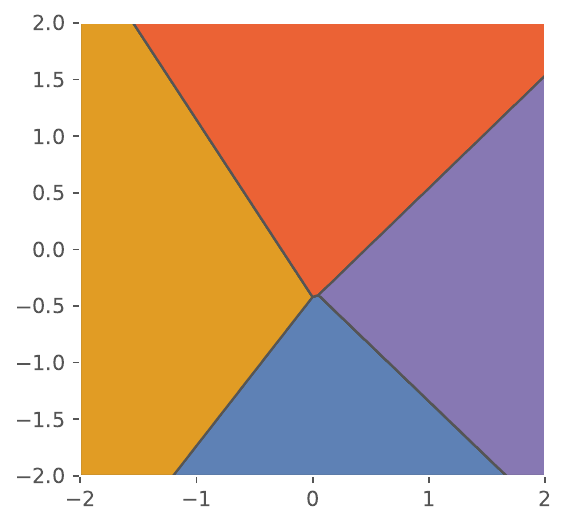}%
&%
\includegraphics[width=0.15\textwidth]{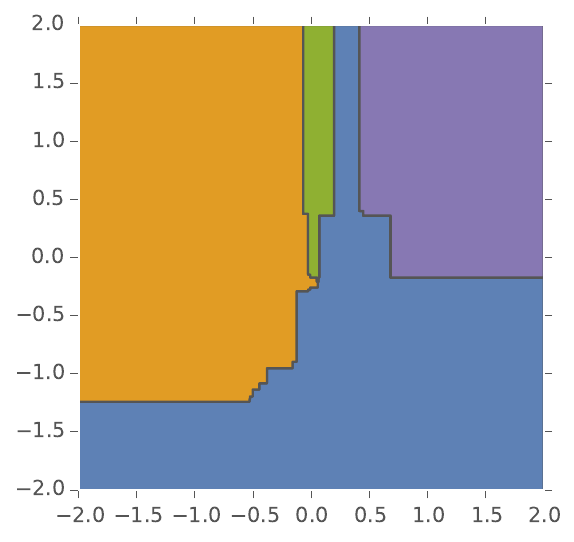}%
\\%
~~~~DR\hfill.26~&~~~~Grad Boost\hfill0.20~&~~~~DR-SVM\hfill0.18~&~~~~PF\hfill0.23~%
\end{tabular}%
\egroup
\end{figure}%
\begin{theorem}\label{consistencythm}
Fix $\pi$ and
let $W_n^*(\pi)=W_n^*(\pi;\fmagd f_{p,\mathcal K_{1:m},\gamma_{n,1:m}},\Lambda_n)$ 
with $0\prec\underline\kappa I\preceq\Lambda_n\preceq\overline\kappa I$,
$0<\underline\gamma\leq\gamma_{n,t}\leq\overline\gamma\,\forall t\in[m]$ for each $n$.
Suppose Asns.~\ref{unconfoundedness} and \ref{weakoverlap} 
hold, $\op{Var}(Y\mid X)$ a.s. bounded,
$\Efb{\sqrt{\mathcal K_t(X,X)}}<\infty$, and
$\Efb{\mathcal K_t(X,X){{\pi^2_T(X)}/{\logging^2_T(X)}}}<\infty$. %
Then the following two results hold:
\begin{enumerate}[label=(\alph*), align=left, leftmargin=*,labelindent=0in,topsep=-1ex,itemsep=0ex,partopsep=0ex,parsep=0ex]
\item\label{consistencythm:finitenorm} {If $\fmagd{\mu_t}_{\mathcal K_t}<\infty$ for all $t\in[m]$:}\hfill${\hat\tau_{W_n^*(\pi)}-\op{SAPE}(\pi)}=O_p(1/\sqrt{n}).$
\item\label{consistencythm:univ} {If $\mathcal K_t$ is $C_0$-universal for all $t\in[m]$:}\hfill${\hat\tau_{W_n^*(\pi)}-\op{SAPE}(\pi)}=o_p(1)$.
\end{enumerate}
\end{theorem}
The key assumptions of Thm.~\ref{consistencythm} are unconfoundedness, overlap, and bounded variance. The other conditions simply guide the choice of method parameters. The two conditions on the kernel are trivial for bounded kernels like the RBF kernel.
An analogous result for the DR estimator is a corollary.
\begin{corollary}\label{consistencycor}
Suppose the assumptions of Thm.~\ref{consistencythm} hold and.
Then
\begin{enumerate}[label=(\alph*), align=left, leftmargin=*,labelindent=0in,topsep=-1ex,itemsep=0ex,partopsep=0ex,parsep=0ex]
\item\label{consistencycor:fast} {If $\fmagd{\hat\mu_{nt}-\mu_t}_{\mathcal K_t}=o_p(1)\,\forall t\in[m]$:}\\\phantom{.}\hfill{${\hat\tau_{W_n^*(\pi),\hat\mu_n}-\op{SAPE}(\pi)}=\fprns{\frac1{n^2}\sum_{i=1}^n{W^*_{ni}}^2\op{Var}(Y_i\mid X_i)}^{1/2}+o_p(1/\sqrt{n}).$}
\item\label{consistencycor:specified} If $\fmagd{\hat\mu_{n}(X)-\mu(X)}_2=O_p(r(n))$, $r(n)=\Omega(1/\sqrt{n})$:\hfill{${\hat\tau_{W_n^*(\pi),\hat\mu_n}-\op{SAPE}(\pi)}=O_p(r(n)).$}
\item\label{consistencycor:finitenorm} {If $\fmagd{\mu_t}_{\mathcal K_t}<\infty,\fmagd{\hat\mu_{nt}}_{\mathcal K_t}=O_p(1)$ for all $t\in[m]$:}\hfill${\hat\tau_{W_n^*(\pi),\hat\mu_n}-\op{SAPE}(\pi)}=O_p(1/\sqrt{n}).$
\item\label{consistencycor:univ} {If $\mathcal K_t$ is $C_0$-universal for all $t\in[m]$:}
\hfill${\hat\tau_{W_n^*(\pi),\hat\mu_n}-\op{SAPE}(\pi)}=o_p(1)$.
\end{enumerate}
\end{corollary}

Cor.~\ref{consistencycor}\ref{consistencycor:fast} is the case where both the balancing
weights and the regression function are well-specified, in which case 
the multiplicative bias disappears faster than $o_p(1/\sqrt{n})$, leaving us only 
with the irreducible residual variance, leading to an \emph{efficient} evaluation.
The other cases concern the ``doubly robust'' nature of the balanced DR estimator:
Cor.~\ref{consistencycor}\ref{consistencycor:specified} requires only that the regression be consitent and Cor.~\ref{consistencycor}\ref{consistencycor:finitenorm}-\ref{consistencycor:univ} require only the balancing weights to be consistent.

\section{Balanced Learning}

Next we consider a balanced approach to policy learning.
Given a policy class $\Pi\subset[\mathcal X\to\Delta^m]$,
we let the \textbf{balanced policy learner} yield the policy $\pi\in\Pi$
that minimizes the balanced policy evaluation using either a vanilla or DR
estimator plus
a potential regularization term in the worst-case/posterior CMSE of the evaluation. 
We formulate this as a \emph{bilevel} optimization
problem:
\begin{talign}\ts\label{balancedlearning}
\hat\pi^\text{bal}
&\in\argmin_\pi
\fbraces{\hat\tau_{W}\hspace{0.385em}+\hspace{-0.2em}\lambda\mathfrak E(W,\pi;\fmagd\cdot,\Lambda)\hspace{-.05em}:\hspace{-.05em}\pi\in\Pi,
W\in\argmin_{W\in\mathcal W}\mathfrak E^2(W,\pi;\fmagd\cdot,\Lambda)}
\tendl\ts\label{DRbalancedlearning}
\hat\pi^\text{bal-DR}
&\in\argmin_\pi
\fbraces{\hat\tau_{W,\hat \mu}\hspace{-0.2em}+\hspace{-0.2em}\lambda\mathfrak E(W,\pi;\fmagd\cdot,\Lambda)\hspace{-.05em}:\hspace{-.05em}\pi\in\Pi,
W\in\argmin_{W\in\mathcal W}\mathfrak E^2(W,\pi;\fmagd\cdot,\Lambda)}
\end{talign}
The regularization term regularizes \emph{both} the \emph{balance}
(\ie, worst-case/posterior bias) that is achievable for $\pi$ 
\emph{and} the \emph{variance} in evaluating $\pi$.
We include this regularizer for completeness and
motivated by the results of 
\citep{swaminathan2015counterfactual} (which regularize variance),
but find that it not necessary to include it in practice.

\subsection{Optimizing the Balanced Policy Learner}

Unlike \citep{beygelzimer2009offset,dudik2011doubly,athey2017efficient,strehl2010learning,zhou2017residual}, our (nonconvex) policy
optimization problem does \emph{not} reduce to weighted classification
precisely because our weights are \emph{not} multiplies of $\pi_{T_i}(X_i)$
(but therefore our weights also lead to better performance).
Instead, like \citep{swaminathan2015counterfactual}, 
we use gradient descent. For that, we need to be able to differentiate
our bilevel optimization problem.
We focus on $p=2$ for brevity.

\begin{theorem}\label{derivtheorem}
Let $\fmagd\cdot=\fmagd\cdot_{2,\mathcal K_{1:m},\gamma_{1:m}}$.
Then $\exists W^*(\pi)\in\argmin_{W\in\mathcal W}\mathfrak E^2(W,\pi;\fmagd\cdot,\Lambda)$ such that
{\footnotesize\begin{talign*}
&\nabla_{\pi_t(X_1),\dots,\pi_t(X_n)} \hat\tau_{W^*(\pi)}=
\frac1n Y_{1:n}^T\tilde H (I - (A+(I-A)\tilde H)^{-1}(I-A)\tilde H)J_t\\
&\nabla_{\pi_t(X_1),\dots,\pi_t(X_n)} \hat\tau_{W^*(\pi),\hat\mu}=
\frac1n\hat\epsilon_{1:n}^T\tilde H (I - (A+(I-A)\tilde H)^{-1}(I-A)\tilde H)J_t+\frac1n\hat\mu_t(\Xs)\\
&\nabla_{\pi_t(X_1),\dots,\pi_t(X_n)} \mathfrak E(W^*(\pi),\pi;\fmagd\cdot,\Lambda)=-D_t/\mathfrak E(W^*(\pi),\pi;\fmagd\cdot,\Lambda)
\end{talign*}}
where
$\tilde H=-F(F^THF)^{-1}F^T$, $F_{ij}=\delta_{ij}-\delta_{in}$ for $i\in[n],j\in[n-1]$,
$A_{ij}=\delta_{ij}\indic{W_i^*(\pi)>0}$,
$D_{ti}=\gamma_t^2\sum_{j=1}^n\mathcal K_t(X_i,X_j)(W_j\delta_{T_jt}-\pi_t(X_j))$,
$H_{ij}=2\sum_{t=1}^m\gamma_t^2\delta_{T_it}\delta_{T_jt}\mathcal K_t(X_i,X_j)+2\Lambda$,
and $J_{tij}=-2\gamma_t^2\delta_{T_it}\mathcal K_t(X_i,X_j)$.
\end{theorem}
To leverage this result, we use a parameterized policy class such as
$\Pi_\text{logit}=\fbraces{\pi_t(x;\beta_t)\propto \exp(\beta_{t0}+\beta_t^Tx)}$ 
(or kernelized versions thereof), 
apply chain rule to differentiate objective in the parameters $\beta$, 
and use BFGS \citep{fletcher2013practical} with random starts.
The logistic parametrization allows us to smooth the problem even while
the solution ends up being deterministic (extreme $\beta$).

This approach requires solving a quadratic program for each objective gradient evaluation.
While this can be made faster by using the previous solution as warm start,
it is still computationally intensive, especially as the bilevel problem is nonconvex
and both it and each quadratic program solved in ``batch'' mode. This is a limitation
of the current optimization algorithm that we hope to improve on in the future
using specialized methods for bilevel optimization \citep{ochs2016techniques,bennett2008bilevel,sabach2017first}.

\begin{example}\label{ex2}
We return to Ex.~\ref{ex1} and consider policy learning.
We use the fixed draw shown in Fig.~\ref{figex1xplot}
and set $\sigma$ to 0. 
We consider a variety of policy learners and plot the policies in
Fig.~\ref{figex2} along with their population regret
$\op{PAPE}(\pi)-\op{PAPE}(\pi^*)$.
The policy learners we consider are:
minimizing standard IPW and DR evaluations over $\Pi_\text{logit}$ with
$\hat\logging,\hat\mu$ as in Ex.~\ref{ex1}
(versions with combinations of normalized, clipped, and/or true $\logging$, 
not shown, all have regret 0.26--0.5),
the direct method with Gaussian 
process regression gradient boosted trees (both sklearn defaults),
weighted SVM classification using IPW and DR weights (details in supplement),
SNPOEM \citep{swaminathan2015self}, PF \citep{kallus2016recursive},
and our balanced policy learner \eqref{balancedlearning}
with parameters as in Ex.~\ref{ex1},
$\Pi=\Pi_\text{logit},\lambda=\Lambda=0$ (the DR version
\eqref{DRbalancedlearning}, not shown, has regret .08).
\end{example}

\begin{example}\label{ex3}
Next, we consider two UCI multi-class classification datasets
\citep{UCI},
Glass ($n=214$, $d=9$, $m=6$) and Ecoli ($n=336$, $d=7$, $m=8$), 
and use a supervised-to-contextual-bandit transformation \citep{beygelzimer2009offset,swaminathan2015counterfactual,dudik2011doubly}
to compare different policy learning algorithms.
Given a supervised multi-class dataset, we
draw $T$ as per a multilogit model with random $\pm1$ coefficients
in the normalized covariates $X$. Further,
we set $Y$ to 0 if $T$ matches the label and 1 otherwise. 
And we split the data 75-25 into training and test sample. 
Using 100 replications of this process, we evaluate the performance
of learned \emph{linear} policies by comparing 
the linear policy learners
as in Ex.~\ref{ex2}.  
For IPW-based approaches, we estimate $\hat\logging$
by a multilogit regression (well-specified by construction).
For DR approaches, we estimate
$\hat\mu$ using gradient boosting trees (sklearn defaults).
We compare these to our balanced policy learner in
both vanilla and DR forms with all parameters fit by
marginal likelihood using the RBF kernel with
an unspecified length scale after normalizing the data.
We tabulate the results in Tab.~\ref{tabex3}.
They first demonstrate that employing the various stopgap fixes to
IPW-based policy learning as in SNPOEM indeed provides a critical
edge. 
This is further improved upon
by using a balanced approach to policy learning, which gives the
best results. 
In this example, DR approaches do worse than vanilla ones, 
suggesting both that XGBoost provided a bad outcome model and/or
that the additional variance of DR was not compensated for by 
sufficiently less bias.
\end{example}
\begin{table*}[t!]\centering\footnotesize
\setlength{\tabcolsep}{.7em}%
\caption{Policy learning results in Ex.~\ref{ex3}}\label{tabex3}
\begin{tabular}{m{1cm}cccccccc}\toprule
&IPW&DR&IPW-SVM&DR-SVM&POEM&SNPOEM&Balanced&Balanced-DR\\\midrule
Glass
&
0.726&
0.755&
0.641&
0.731&
0.851&
0.615&
\textbf{0.584}&
0.660
\\
Ecoli
&
0.488&
0.501&
0.332&
0.509&
0.431&
0.331&
\textbf{0.298}&
0.371
\\\bottomrule
\end{tabular}
\end{table*}

\subsection{Uniform Consistency and Regret Bounds}

Next, we establish consistency results \emph{uniformly} over policy classes.
This allows us to bound the regret of the balanced policy learner.
We define the sample and population regret, respectively, as
$$\ts R_\Pi(\hat\pi)=\op{PAPE}(\hat\pi)-\min_{\pi\in\Pi}\op{PAPE}(\pi),\quad
\widehat R_\Pi(\hat\pi)=\op{SAPE}(\hat\pi)-\min_{\pi\in\Pi}\op{SAPE}(\pi)$$
A key requirement for these to converge is that the best-in-class policy
is learnable. We quantify that using Rademacher complexity \citep{bartlett2003rademacher} and later extend
our results to VC dimension. Let us define
$$\ts\widehat{\mathfrak R}_n(\mathcal F)=\frac1{2^n}\sum_{\rho_i\in\fbraces{-1,+1}^n}\sup_{f\in\mathcal F}\frac1n\sum_{i=1}^n\rho_if(X_i),\quad
{\mathfrak R}_n(\mathcal F)=\Efb{\widehat{\mathfrak R}_n(\mathcal F)}.$$
\emph{E.g.}, for linear policies $\widehat{\mathfrak R}_n(\mathcal F)=O(1/\sqrt{n})$ \citep{kakade2009complexity}.
If $\mathcal F\subseteq[\mathcal X\to\Rl^m]$ let $\mathcal F_t=\fbraces{(f(\cdot))_t:f\in\mathcal F}$ and set ${\mathfrak R}_n(\mathcal F)=\sum_{t=1}^m{\mathfrak R}_n(\mathcal F_t)$ and same for $\widehat{\mathfrak R}_n(\mathcal F)$.
We also strengthen the overlap assumption.
\begin{assumption}[Strong overlap]\label{strongoverlap}
$\exists\alpha\geq1$ such that $\fPrb{\logging_t(X)\geq1/\alpha}=1$ $\forall t\in[m]$. 
\end{assumption}
\begin{theorem}\label{uniformconsistencythm}
Fix $\Pi\subseteq[\mathcal X\to\Delta^m]$ and
let $W_n^*(\pi)=W_n^*(\pi;\fmagd f_{p,\mathcal K_{1:m},\gamma_{n,1:m}},\Lambda_n)$ 
with $0\prec\underline\kappa I\preceq\Lambda_n\preceq\overline\kappa I$,
$0<\underline\gamma\leq\gamma_{n,t}\leq\overline\gamma\,\forall t\in[m]$
for each $n$ and $\pi\in\Pi$.
Suppose Asns.~\ref{unconfoundedness} and \ref{strongoverlap} hold, 
$\abs{\epsilon_i}\leq B$ a.s. bounded, and
$\sqrt{\mathcal K_t(x,x)}\leq \Gamma\,\forall t\in[m]$ for $\Gamma\geq1$.
Then the following two results hold:
\begin{enumerate}[label=(\alph*), align=left, leftmargin=*,labelindent=0in,topsep=-1ex,itemsep=0ex,partopsep=0ex,parsep=0ex]
\item\label{uniformconsistencythm:finitenorm} If $\fmagd{\mu_t}_{\mathcal K_t}<\infty,\,\forall t\in[m]$ then
for $n$ sufficiently large ($n\geq 2\log(4m/\nu)/(1/(2\alpha)-{\mathfrak R}_n(\Pi))^2$), we have that, with probability at least $1-\nu$,
\begin{talign*}
\hspace{-2.1em}\sup_{\pi\in\Pi}
\fabs{\hat\tau_{W^*(\pi)}-\op{SAPE}(\pi)}\leq&
8\alpha\Gamma\overline\gamma m(\fmagd\mu+\sqrt{2\log(4m/\nu)}\underline\kappa^{-1}B){\mathfrak R}_n(\Pi)
\tendl&+\frac{1}{\sqrt n}\prns{{2\alpha\overline\kappa\fmagd\mu+12\alpha\Gamma^2\overline\gamma m\fmagd\mu}+6\alpha\Gamma\overline\gamma m\underline\kappa^{-1}B\log\prns{\frac{4m}{\nu}}}
\tendl&+\frac{1}{\sqrt n}\fprns{
2\alpha\overline\kappa\underline\kappa^{-1}B+
12\alpha\Gamma^2\overline\gamma m\underline\kappa^{-1}B+
3\alpha\Gamma\overline\gamma m\fmagd\mu}\sqrt{2\log\prns{\frac{4m}{\nu}}}
\end{talign*}
\item\label{uniformconsistencythm:univ} If $\mathcal K_t$ is $C_0$-universal for all $t\in[m]$ and either $\ts{\mathfrak R}_n(\Pi)=o(1)$ or $\ts\widehat{\mathfrak R}_n(\Pi)=o_p(1)$ then $$\ts\sup_{\pi\in\Pi}\fabs{\hat\tau_{W^*(\pi)}-\op{SAPE}(\pi)}=o_p(1).$$
\end{enumerate}
\end{theorem}
The proof crucially depends on \emph{simultaneously} handling
the functional complexities of both the policy class $\Pi$ \emph{and} 
the space of functions $\fbraces{f:\fmagd f<\infty}$ being
balanced against.
Again, the key assumptions of Thm.~\ref{uniformconsistencythm} are unconfoundedness, overlap, and bounded residuals. The other conditions simply guide the choice of method parameters. 
Regret bounds follow as a corollary.
\begin{corollary}\label{uniformconsistencycor1}
Suppose the assumptions of Thm.~\ref{uniformconsistencythm} hold. 
If $\hat\pi_n^\text{bal}$ is as in \eqref{balancedlearning} then:
\begin{enumerate}[label=(\alph*), align=left, leftmargin=*,labelindent=0in,topsep=-1ex,itemsep=0ex,partopsep=0ex,parsep=0ex]
\item\label{vanillaregret:finitenorm} If 
$\fmagd{\mu_t}_{\mathcal K_t}<\infty$ for all $t\in[m]$:\hfill%
$\ts
R_\Pi(\hat\pi_n^\text{bal})= O_p(\mathfrak R_n(\Pi)+1/\sqrt{n}).
$
\item\label{vanillaregret:univ} If 
$\mathcal K_t$ is $C_0$-universal for all $t\in[m]$:\hfill%
$\ts
R_\Pi(\hat\pi_n^\text{bal})=o_p(1).
$
\end{enumerate}
If $\hat\pi_n^\text{bal-DR}$ is as in \eqref{DRbalancedlearning} then:
\begin{enumerate}[label=(\alph*), align=left, leftmargin=*,labelindent=0in,topsep=-1ex,itemsep=0ex,partopsep=0ex,parsep=0ex]
\setcounter{enumi}{2}
\item\label{DRregret:fast} If 
$\fmagd{\hat\mu_{nt}-\mu_t}_{\mathcal K_t}=o_p(1)$ 
for all $t\in[m]$:\hfill%
$\ts
R_\Pi(\hat\pi_n^\text{bal-DR})= O_p(\mathfrak R_n(\Pi)+1/\sqrt{n}).
$
\item\label{DRregret:specified} If 
$\fmagd{\hat\mu_n(X)-\mu(X)}_2=O_p(r(n))$:\hfill%
$\ts
R_\Pi(\hat\pi_n^\text{bal-DR})= O_p(r(n)+\mathfrak R_n(\Pi)+1/\sqrt{n}).
$
\item\label{DRregret:finitenorm} If 
$\fmagd{\mu_t}_{\mathcal K_t}<\infty,\fmagd{\hat\mu_{nt}}_{\mathcal K_t}=O_p(1)$ 
for all $t\in[m]$:\hfill%
$\ts
R_\Pi(\hat\pi_n^\text{bal-DR})= O_p(\mathfrak R_n(\Pi)+1/\sqrt{n}).
$
\item\label{DRregret:univ} If 
$\mathcal K_t$ is $C_0$-universal for all $t\in[m]$:\hfill%
$\ts
R_\Pi(\hat\pi_n^\text{bal})=o_p(1).
$
\end{enumerate}
And, all the same results hold when replacing ${\mathfrak R}_n(\Pi)$ with $\widehat{\mathfrak R}_n(\Pi)$ and/or replacing $R_\Pi$ with $\widehat R_\Pi$.
\end{corollary}
\section{Conclusion}

Considering the policy evaluation and learning problems
using observational or logged data, we presented a new method
that is based on finding optimal balancing weights that make the 
data look like the target policy and that is aimed at ameliorating
the shortcomings of existing methods, which included having to
deal with near-zero propensities, using too few positive weights,
and using an awkward two-stage procedure. The new approach showed
promising signs of fixing these issues in some numerical examples.
However, the new learning method is more computationally intensive
than existing approaches, solving a QP at each gradient step.
Therefore, in future work, we plan to explore faster algorithms that can implement
the balanced policy learner, perhaps using alternating descent,
and use these to investigate comparative numerics in much larger datasets.

\clearpage

\section*{Acknowledgements} 
This material is based upon work supported by the National Science Foundation under Grant No. 1656996.

\bibliography{balpol}
\bibliographystyle{abbrvnat}

\clearpage

\appendix

\section{Omitted Proofs}

\begin{proof}[Proof of Thm.~\ref{cmsethm}]
Noting that $Y_i=Y_i(T_i)=\sum_{t=1}^m\delta_{T_it}\mu_t(X_i)+\epsilon_{i}$, let us rewrite $\hat\tau_W$ as
$$\ts
\hat\tau_W=\frac1n\sum_{t=1}^m\sum_{i=1}^nW_i\delta_{T_it}\mu_t(X_i)+\frac1n\sum_{i=1}^nW_i\epsilon_{i}.
$$
Recalling that $\op{SAPE}(\pi)=\frac1n\sum_{i=1}^n\sum_{t=1}^m\pi_t(X_i)\mu_t(X_i)$ 
immediately yields the first result. To obtain the second result note that $\op{SAPE}(\pi)$ 
is measurable with respect to $X_{1:n},T_{1:n}$ so that
$$\ts
\op{CMSE}(\hat\tau_W,\pi)=
\fprns{\Efb{\hat\tau_W\mid X_{1:n},T_{1:n}}-\op{SAPE}(\pi)}^2
+
\op{Var}\fprns{\hat\tau_W\mid X_{1:n},T_{1:n}}.
$$
By Asn.~\ref{unconfoundedness}
$$\ts
\Efb{\delta_{T_it}\epsilon_{i}\mid X_{1:n},T_{1:n}}=
\delta_{T_it}\Efb{\epsilon_{i}\mid X_i}=
\delta_{T_it}(\Efb{Y_i(t)\mid X_i}-\mu_t(X_i))=0.
$$
Therefore, $$\ts\Efb{\hat\tau_W\mid X_{1:n},T_{1:n}}=\frac1n\sum_{t=1}^m\sum_{i=1}^nW_i\delta_{T_it}\mu_t(X_i),$$ giving the first term of $\op{CMSE}(\hat\tau_W,\pi)$.
Moreover, since $$\ts\Efb{\epsilon_{i}\epsilon_{i'}\mid X_{1:n},T_{1:n}}=\delta_{ii'}\sigma_{T_i}^2,$$ we have
\begin{talign*}
\op{Var}\fprns{\hat\tau_W\mid X_{1:n},T_{1:n}}&=
\Efb{\fprns{\hat\tau_W-\Efb{\hat\tau_W\mid X_{1:n},T_{1:n}}}^2\mid X_{1:n},T_{1:n}}
\tendl&=
\frac1{n^2}\Efb{\fprns{\sum_{i=1}^nW_i\epsilon_{i}}^2\mid X_{1:n},T_{1:n}}=\frac1{n^2}\sum_{i=1}^nW_i^2\sigma_{T_i}^2,
\end{talign*}
giving the second term.
\end{proof}

\begin{proof}[Proof of Cor.~\ref{drcor}]
This follows from Thm.~\ref{cmsethm} after noting 
that 
$\ts\hat\tau_{W,\hat\mu}=\hat\tau_{W}-B(W,\pi;\hat\mu)$
and that $B_t(W,\pi;\hat\mu_t)-B_t(W,\pi;\hat\mu_t)=B_t(W,\pi;\mu_t-\hat\mu_t)$.
\end{proof}

\begin{proof}[Proof of Lemma~\ref{boundlemma}]
For the first statement, we have
\begin{talign*}
\mathfrak E^2(W,\pi;\fmagd \cdot_{p,\mathcal K_{1:m},\gamma_{1:m}},\Lambda)
&=
\sup_{\fmagd v_p\leq1,
\fmagd{f_t}_{\mathcal K_t}\leq \gamma_t v_t}\fprns{
\sum_{t=1}^mB_t(W,\pi_t;f_t)
}^2+\frac1{n^2}W^T\Lambda W
\\&=
\sup_{\fmagd v_p\leq1}\fprns{
\sum_{t=1}^m\sup_{\fmagd{f_t}_{\mathcal K_t}\leq \gamma_t v_t}B_t(W,\pi_t;f_t)
}^2+\frac1{n^2}W^T\Lambda W
\\&=
\sup_{\fmagd v_p\leq1}\fprns{
\sum_{t=1}^mv_t \gamma_t \mathfrak B_t(W,\pi_t;\fmagd\cdot_{\mathcal K_t})
}^2+\frac1{n^2}W^T\Lambda W
\\&=
\fprns{\sum_{t=1}^m\gamma_t^q\mathfrak B_t^q(W,\pi_t;\fmagd\cdot_{\mathcal K_t})}^{2/q}
+\frac1{n^2}W^T\Lambda W.
\end{talign*}

For the second statement, 
let $z_{ti}=(W_i\delta_{T_it}-\pi_t(X_i))$ and note that
since $\Efb{(\mu_t(X_i)-f_t(X_i))(\mu_s(X_j)-f_s(X_j))\mid\XTs}=\delta_{ts}\mathcal K_t(X_i,X_j)$, 
we have
\begin{talign*}
\op{CMSE}(\hat\tau_{W,f},\pi)&=
\Efb{
(\sum_{t=1}^m B_t(W,\pi_t;\mu_t-f_t))^2\mid\XTs
}+\frac{1}{n^2}W^T\Sigma W
\\&=
\sum_{t,s=1}^m\sum_{i,j=1}^m
z_{ti}z_{sj}\Efb{(\mu_t(X_i)-f_t(X_i))(\mu_s(X_j)-f_s(X_j))
\mid\XTs}\\&\phantom{=}+\frac{1}{n^2}W^T\Sigma W
\\&=
\sum_{t=1}^m\sum_{i,j=1}^mz_{ti}z_{tj}\mathcal K_t(X_i,X_j)+\frac{1}{n^2}W^T\Sigma W.
\end{talign*}
\end{proof}

\begin{proof}[Proof of Thm.~\ref{consistencythm}]
Let $Z=\frac1n\sum_{i=1}^n\pi_{T_{i}}(X_{i})/\logging_{T_{i}}(X_{i})$ and $\tilde W_i(\pi)=\frac1{Z}{\pi_{T_i}(X_i)/\logging_{T_i}(X_i)}$ and note that $\tilde W\in\mathcal W$.
Moreover, note that
\begin{talign*}
\mathfrak B_t(\tilde W,\pi_t;\fmagd\cdot_{\mathcal K_t})&=
\frac{1}{Z}\fmagd{\frac1n\sum_{i=1}^n\fprns{\frac{\delta_{T_it}}{\logging_t(X_i)}-Z}\pi_t(X_i)E_{X_i}}_{\mathcal K_t}
\tendl&\leq
\frac{1}{Z}\fmagd{\frac1n\sum_{i=1}^n\fprns{\frac{\delta_{T_it}}{\logging_t(X_i)}-1}\pi_t(X_i)E_{X_i}}_{\mathcal K_t}
+\frac{1}{Z}\fmagd{\frac1n\sum_{i=1}^n\fprns{Z-1}\pi_t(X_i)E_{X_i}}_{\mathcal K_t}
\tendl&\leq
\frac{1}{Z}\fmagd{\frac1n\sum_{i=1}^n\fprns{\frac{\delta_{T_it}}{\logging_t(X_i)}-1}\pi_t(X_i)E_{X_i}}_{\mathcal K_t}
+\frac{\fabs{Z-1}}{Z}
\frac1n\sum_{i=1}^n{\sqrt{\mathcal K_t(X_i,X_i)}}.
\end{talign*}
Let $\xi_i=\fprns{\frac{\delta_{T_it}}{\logging_t(X_i)}-1}\pi_t(X_i)E_{X_i}$ and note that
$\Efb{\xi_i}=\Efb{\fprns{\Efb{\delta_{T_it}/\logging_t(X_i)\mid X_i}-1}\pi_t(X_i)E_{X_i}}=0$ and that $\xi_1,\xi_2,\dots$ are iid. Therefore, letting $\xi_1',\xi_2',\dots$  be iid replicates of $\xi_1,\xi_2,\dots$ (ghost sample) and letting $\rho_i$ be iid Rademacher random variables independent of all else, we have
\begin{talign*}
\Efb{\fmagd{\frac1n\sum_{i=1}^n\xi_i}_{\mathcal K_t}^2}
&=
\frac1{n^2}\Efb{\fmagd{\sum_{i=1}^n(\Efb{\xi'_i}-\xi_i)}_{\mathcal K_t}^2}
\leq
\frac1{n^2}\Efb{\fmagd{\sum_{i=1}^n(\xi'_i-\xi_i)}_{\mathcal K_t}^2}
\tendl&=
\frac1{n^2}\Efb{\fmagd{\sum_{i=1}^n\rho_i(\xi'_i-\xi_i)}_{\mathcal K_t}^2}
\leq
\frac4{n^2}\Efb{\fmagd{\sum_{i=1}^n\rho_i\xi_i}_{\mathcal K_t}^2}
\end{talign*}
Note that $\fmagd{\xi_1-\xi_2}_{\mathcal K_t}^2+\fmagd{\xi_1+\xi_2}^2_{\mathcal K_t}=2\fmagd{\xi_1}^2_{\mathcal K_t}+2\fmagd{\xi_2}^2_{\mathcal K_t}+2\ip{\xi_1}{\xi_2}-2\ip{\xi_1}{\xi_2}=2\fmagd{\xi_1}^2_{\mathcal K_t}+2\fmagd{\xi_2}^2_{\mathcal K_t}$.
By induction, $\sum_{\rho_i\in\{-1,+1\}^n}\fmagd{\sum_{i=1}^n\rho_i\xi_i}_{\mathcal K_t}^2=2^n\sum_{i=1}^n\fmagd{\xi_i}_{\mathcal K_t}^2$. Since 
\begin{talign*}
\Efb{\fmagd{\xi_i}_{\mathcal K_t}^2}
&\leq
2\Efb{{\frac{\pi^2_T(X)}{\logging^2_T(X)}}\mathcal K_t(X,X)}+2\Efb{\pi^2_t(X)\mathcal K_t(X,X)}
\leq
4\Efb{{\frac{\pi^2_T(X)}{\logging^2_T(X)}}
\mathcal K_t(X,X)
}
<\infty,
\end{talign*}
we get
$\ts
\Efb{\fmagd{\frac1n\sum_{i=1}^n\xi_i}_{\mathcal K_t}^2}=O(1/n)$
and therefore $\fmagd{\frac1n\sum_{i=1}^n\xi_i}_{\mathcal K_t}^2=O_p(1/n)$
by Markov's inequality.
Moreover, as $\Efb{\pi_{T}(X)/\logging_{T}(X)}=\Efb{\sum_{t=1}^m\Efb{\delta_{Tt}\mid X}\pi_t(X)/\logging_t(X)}=\Efb{\sum_{t=1}^m\pi_t(X)}=1$ and $\Efb{\pi^2_{T}(X)/\logging_{T}(X)^2}<\infty$, by Chebyshev's inequality, $\Efb{\fprns{Z-1}^2}=O(1/n)$ so that
$(Z-1)^2=O_p(1/n)$ by Markov's inequality.
Similarly, as $\Efb{\sqrt{\mathcal K_t(X,X)}}<\infty$, we have
$\frac1n\sum_{i=1}^n{\sqrt{\mathcal K_t(X_i,X_i)}}\to_p\Efb{\sqrt{\mathcal K_t(X,X)}}$. Putting it all together, by Slutsky's theorem, $\mathfrak B^2_t(\tilde W,\pi_t;\fmagd\cdot_{\mathcal K_t})=O_p(1/n)$. Moreover, $\fmagd{\tilde W}^2_2=\frac1{Z^2}\sum_{i=1}^n\pi^2_{T_i}(X_i)/\logging^2_{T_i}(X_i)=O_p(n)$.
Therefore, since $\Lambda_n\preceq \overline\kappa I$ and since $W_n^*$ is optimal and $\tilde W\in\mathcal W$, we have
\begin{talign*}
\mathfrak E^2(W_n^*,\pi;\fmagd \cdot_{p,\mathcal K_{1:m},\gamma_{n,1:m}},\Lambda_n)&\leq\mathfrak E^2(\tilde W,\pi;\magd \cdot_{p,\mathcal K_{1:m},\gamma_{n,1:m}},\Lambda_n)
\tendl&\leq 
\overline\gamma^2\prns{\sum_{t=1}^m\mathfrak B^q_t(\tilde W,\pi_t;\fmagd\cdot_{\mathcal K_t})}^{2/q}+\frac{\overline\kappa}{n^2}\fmagd{\tilde W}^2_2=O_p(1/n)
\end{talign*}
Therefore,
\begin{talign*}\ts
\mathfrak B^2_t(W_n^*,\pi_t;\fmagd\cdot_{\mathcal K_t})\leq \underline\gamma^{-2}\mathfrak E^2(W_n^*,\pi;\fmagd\cdot_{1:m},\gamma_{n,1:m},\Lambda_n)&=O_p(1/n),\tendl
\frac1{n^2}\fmagd{W_n^*}_2^2\leq \frac1{\underline\kappa n^2}{W_n^*}^T\Lambda_n {W_n^*}\leq\underline\kappa^{-1}\mathfrak E^2(W_n^*,\pi;\fmagd\cdot_{1:m},\gamma_{n,1:m},\Lambda_n)&=O_p(1/n).
\end{talign*}

Now consider case \ref{consistencythm:finitenorm}. 
By assumption $\fmagd\Sigma_2\leq\overline\sigma^2<\infty$ for all $n$.
Then we have
\begin{talign*}\ts
\op{CMSE}(\hat\tau_{W_n^*},\pi)
&\leq
m\sum_{t=1}^m \fmagd{\mu_t}^2_{\mathcal K_t}\mathfrak B^2_t(W_n^*,\pi_t;\fmagd\cdot_{\mathcal K_t})+\frac{\overline\sigma^2}{n^2}\fmagd{W_n^*}_2^2=O_p(1/n).
\end{talign*}
Letting $D_n=\sqrt{n}\abs{\hat\tau_{W_n^*}-\op{SAPE}(\pi)}$ and
$\mathcal G$ be the sigma algebra of $X_1,T_1,X_2,T_2,\dots$, 
Jensen's inequality yields $\Efb{D_n\mid \mathcal G}=O_p(1)$ from the above.
We proceed to show that $D_n=O_p(1)$, yielding the first result.
Let $\nu>0$ be given. Then
$\Efb{D_n\mid \mathcal G}=O_p(1)$
says that 
there exist $N,M$ such that 
$\fPrb{\Efb{D_n\mid \mathcal G}>M}
\leq\nu/2$ for all $n\geq N$.
Let $M_0=\max\{M,2/\nu\}$ and observe that, for all $n\geq N$,
\begin{talign*}
\fPrb{D_n>M_0^2}
&=
\fPrb{D_n>M_0^2,\Efb{D_n\mid \mathcal G}>M_0}
+\fPrb{D_n>M_0^2,\Efb{D_n\mid \mathcal G}\leq M_0}
\tendl&=
\fPrb{D_n>M_0^2,\Efb{D_n\mid \mathcal G}>M_0}
+\Efb{\fPrb{D_n>M_0^2\mid \mathcal G}
\indic{\Efb{D_n\mid \mathcal G}\leq M_0}}
\tendl&\leq
\nu/2
+\Efb{\frac{\Efb{D_n\mid \mathcal G}}{M_0^2}
\indic{\Efb{D_n\mid \mathcal G}\leq M_0}}\leq \nu/2+1/M_0\leq\nu
\end{talign*}

Now consider case \ref{consistencythm:univ}. 
We first show that $B_t(W_n^*,\pi_t;\mu_t)=o_p(1)$.
Fix $t\in[m]$ and $\eta>0,\nu>0$. 
Because $\mathfrak B_t(W_n^*,\pi_t;\fmagd\cdot_{\mathcal K_t})=O_p(n^{-1/2})=o_p(n^{-1/4})$
and $\fmagd {W_n^*}_2=O_p(\sqrt{n})$, there are $M,N$ such that for all $n\geq N$ both
$\fPrb{n^{1/4}\mathfrak B_t(W_n^*,\pi_t;\fmagd\cdot_{\mathcal K_t})>\sqrt\eta}\leq\nu/3$
and
$\fPrb{n^{-1/2}\fmagd {W_n^*}_2>M\sqrt\eta}\leq\nu/3$.
Next, fix $\tau=\sqrt{\nu\eta/3}/M$.
By existence of second moment, there is $g_0'=\sum_{i=1}^\ell \beta_iI_{S_i}$ with $(\Eb{(\mu_t(X)-g_0'(X))^2})^{1/2}\leq\tau/2$ where $I_S(x)$ are the simple functions $I_S(x)=\mathbb I\bracks{x\in S}$ for $S$ measurable. Let $i=1,\dots,\ell$.
Let $U_i\supset S_i$ open and $E_i\subseteq S_i$ compact be such that $\Prb{X\in U_i\backslash E_i}\leq\tau^2/(4\ell\abs{\beta_i})^2$. By Urysohn's lemma \citep{royden1988real}, there exists a continuous function $h_i$ with support $C_i\subseteq U_i$ compact, $0\leq h_i\leq1$, and $h_i(x)=1\,\forall x\in E_i$. Therefore,
$\fprns{\Eb{(I_{S_i}(X)-h_i)^2}}^{1/2}=\fprns{\Eb{(I_{S_i}(X)-h_i)^2\mathbb I\bracks{X\in U_i\backslash E_i}}}^{1/2}\leq \fprns{\Prb{X\in U_i\backslash E_i}}^{1/2}\leq \tau/(4\ell\abs{\beta_i})$.
By $C_0$-universality, 
$\exists g_i=\sum_{j=1}^m\alpha_j\mathcal K_t(x_j,\cdot)$ such that $\sup_{x\in\mathcal X}\abs{h_i(x)-g_i(x)}<\tau/(4\ell\abs{\beta_i})$. 
Because $\Eb{(h_i-g_i)^2}\leq \sup_{x\in\mathcal X}\abs{h_i(x)-g_i(x)}^2$, we have $\sqrt{\Eb{(I_{S'}(X)-g_i)^2}}\leq \tau/(2\ell\abs{\beta_i})$. 
Let $\tilde\mu_t=\sum_{i=1}^\ell \beta_ig_i$. Then 
$\fprns{\Eb{(\mu_t(X)-\tilde\mu_t(X))^2}}^{1/2}\leq \tau/2+\sum_{i=1}^\ell\abs{\beta_i} \tau/(2\ell\abs{\beta_i})=\tau$
and $\magd{\tilde\mu_t}_{\mathcal K_t}<\infty$.
Let $\delta_n=\sqrt{\frac1n\sum_{i=1}^n(\mu_t(X_i)-\tilde\mu_t(X_i))^2}$ 
so that $\E\delta_n^2\leq\tau^2$.
Now, because we have
\begin{talign*}
B_t(W_n^*,\pi_t;\mu_t)&=B_t(W_n^*,\pi_t;\tilde\mu_t)+B_t(W_n^*,\pi_t;\mu_t-\tilde\mu_t)
\tendl&\leq 
\fmagd{\tilde\mu_t}_{\mathcal K_t}\mathfrak B_t(W_n^*,\pi_t;\fmagd\cdot_{\mathcal K_t})
+
\sqrt{\frac1n\sum_{i=1}^n(W_{ni}^*\delta_{T_it}-\pi_t(X_i))^2}\delta_n
\tendl&\leq 
\fmagd{\tilde\mu_t}_{\mathcal K_t}\mathfrak B_t(W_n^*,\pi_t;\fmagd\cdot_{\mathcal K_t})
+
\fprns{n^{-1/2}\fmagd {W_n^*}_2+1}\delta_n,
\end{talign*}
letting $N'=\max\{N,2\lceil\fmagd{\tilde\mu_t}_{\mathcal K_t}^4/\eta^2\rceil\}$,
we must then have, for all $n\geq N'$, by union bound and by Markov's inequality, that
\begin{talign*}
\fPrb{
B_t(W_n^*,\pi_t;\mu_t)>\eta
}\leq&
\fPrb{n^{-1/4}\fmagd{\tilde\mu_t}_{\mathcal K_t}>\sqrt\eta}
+
\fPrb{n^{1/4}\mathfrak B_t(W_n^*,\pi_t;\fmagd\cdot_{\mathcal K_t})>\sqrt\eta}
\tendl&+
\fPrb{n^{-1/2}\fmagd {W_n^*}_2>M\sqrt\eta}\leq\nu/3
+
\fPrb{
\delta_n>\sqrt{\eta}/M
}
\tendl\leq& 0+\nu/3+\nu/3+\nu/3=\nu.
\end{talign*}
Following the same logic as in case \ref{consistencythm:finitenorm}, 
we get $\op{CMSE}(\hat\tau_{W_n^*},\pi)=o_p(1)$, so
letting $D_n=\abs{\hat\tau_{W_n^*}-\op{SAPE}(\pi)}$ and
$\mathcal G$ be as before, we have $\Efb{D_n\mid\mathcal G}=o_p(1)$ by Jensen's inequality.
Let $\eta>0,\nu>0$ be given. Let $N$ be such that $\fPrb{\Efb{D_n\mid\mathcal G}>\nu\eta/2}\leq\nu/2$. Then for all $n\geq N$:
\begin{talign*}
\fPrb{D_n>\eta}&=\fPrb{D_n>\eta,\Efb{D_n\mid \mathcal G}>\eta\nu/2}
+\fPrb{D_n>\eta,\Efb{D_n\mid \mathcal G}\leq \eta\nu/2}
\tendl&=
\fPrb{D_n>\eta,\Efb{D_n\mid \mathcal G}>\eta\nu/2}
+\Efb{\fPrb{D_n>\eta\mid \mathcal G}
\indic{\Efb{D_n\mid \mathcal G}\leq \eta\nu/2}}
\tendl&\leq
\nu/2
+\Efb{\frac{\Efb{D_n\mid \mathcal G}}{\eta}
\indic{\Efb{D_n\mid \mathcal G}\leq \eta\nu/2}}\leq \nu/2+\nu/2\leq\nu,
\end{talign*}
showing that $D_n=o_p(1)$ and completing the proof.
\end{proof}

\begin{proof}[Proof of Cor.~\ref{consistencycor}]
Case \ref{consistencycor:fast} follows directly from the proof of Thm.~\ref{consistencythm} noting that the bias term now disappears at rate $o_p(1)O_p(1/\sqrt{n})=o_p(1/\sqrt{n})$.
For Case \ref{consistencycor:specified}, observe that
by Cauchy-Schwartz and 
Slutsky's theorem
$\abs{B_t(W,\pi_t;\mu-\hat\mu_n)}\leq
\fprns{n^{-1/2}\fmagd{W^*_n}_2+1}
\fprns{\frac1n\sum_{i=1}^n(\hat\mu_n(X_i)-\mu(X_i))^2}^{1/2}=O_p(r_n)$.
For cases in cases \ref{consistencycor:finitenorm} and \ref{consistencycor:univ} we treat
$B_t(W,\pi_t;\mu-\hat\mu_n)$ as in the proof of Thm.~\ref{consistencythm} noting
that $\fmagd{\mu_t-\hat\mu_{nt}}_{\mathcal K_t}\leq\fmagd{\mu_t}_{\mathcal K_t}+\fmagd{\hat\mu_{nt}}_{\mathcal K_t}$ and that, in case \ref{consistencycor:finitenorm}, $\fmagd{\hat\mu_{nt}}_{\mathcal K_t}=O_p(1)$ implies
by Markov's inequality that $\fmagd{\hat\mu_{nt}}_{\mathcal K_t}=O_p(1)$. The rest follows as in the proof of Thm.~\ref{consistencythm}.
\end{proof}

\begin{proof}[Proof of Thm.~\ref{derivtheorem}]
First note that because our problem is a quadratic program, the KKT conditions
are necessary and sufficient and we can always choose
an optimizer where strict complementary slackness holds.

Ignore previous definitions of some symbols,
consider any linearly constrained parametric nonlinear optimization problem in standard 
form: 
$z(x)\in\argmin_{y\geq0,By=b}f(x,y)$ where $x\in\R n$, $y\in\R m$, and $b\in\R \ell$.
KKT says there exist $\mu(x)\in\R m,\lambda(x)\in\R l$ such that
(a) $\nabla_y f(x,z(x))=\mu(x) + B^T\lambda(x)$, (b) $Bz(x)=b$, (c) $z(x)\geq0$,
(d) $\mu(x)\geq0$,
and (e) $\mu(x)\odot z(x)=0$, where $\odot$ is the Hadamard product. Suppose strict 
complementary slackness holds in that (f) $\mu(x) + z(x)>0$. By (a), we have that
$$\ts
\nabla_{xy}f(x,z(x))+\nabla_{yy}f(x,z(x))\nabla_xz(x)
=
\nabla_x\mu(x)+B^T\nabla_x\lambda(x),
$$
and hence, letting $H=\nabla_{yy}f(x,z(x))$ and $J=\nabla_{xy}f(x,z(x))$,
$$\ts
\nabla z(x)=H^{-1}(\nabla_x\mu(x)+B^T\nabla\lambda (x)-J).
$$
By (b), we have that
$
B\nabla z(x)=0
$
so that
$$\ts
BH^{-1}\nabla_x\mu(x)+BH^{-1}B^T\nabla\lambda=BH^{-1}J,
$$
and hence if the columns of $F$ form a basis for the null space of $B$ and 
$\tilde H=-F(F^THF)^{-1}F^T$,
$$\ts
\nabla_xz(x)=(H^{-1}B^T(AH^{-1}A^T)^{-1}AH^{-1}-H^{-1})(J-\nabla_x\mu(x))=\tilde H(J-\nabla_x\mu(x)).
$$
By (e), we have that
$$\ts
z_i(x)\nabla_x\mu_i(x)+\mu_i(x)\nabla_xz_i(x)=0,
$$
and then by (f), letting $A=\op{diag}(\indic{z_1(x)>0,\dots,z_m(x)>0})$ we have 
$$\ts
A\nabla_x\mu(x)=0,\quad (I-A)\nabla_x z(x)=0,
$$
and therefore
$$
A\nabla_x\mu(x)-(I-A)\tilde H(J-\nabla_x\mu(x))=0
$$
yielding finally that
$$
\nabla_xz(x)=\tilde H(I-(A+(I-A)\tilde H)^{-1}(I-A)\tilde H)J.
$$

The rest of the theorem is then begotten by applying this result and using chain rule.
\end{proof}

\begin{proof}[Proof of Thm.~\ref{uniformconsistencythm}]
Let $Z(\pi)=\frac1n\sum_{i=1}^n\pi_{T_{i}}(X_{i})/\logging_{T_{i}}(X_{i})$ and $\tilde W_i(\pi)=\frac1{Z(\pi)}{\pi_{T_i}(X_i)/\logging_{T_i}(X_i)}$ and note that $\tilde W\in\mathcal W$.
Moreover, note that
\begin{talign*}
\sup&_{\pi\in\Pi}\mathfrak B_t(\tilde W,\pi_t;\fmagd\cdot_{\mathcal K_t})=
\sup_{\pi\in\Pi,\fmagd{f_t}_{\mathcal K_t}\leq1}\frac{1}{Z(\pi)}{\frac1n\sum_{i=1}^n\fprns{\frac{\delta_{T_it}}{\logging_t(X_i)}-Z(\pi)}\pi_t(X_i)f_t(X_i)}
\tendl&\leq
\fprns{\sup\limits_{\pi\in\Pi}Z(\pi)^{-1}}\sup\limits_{\pi_t\in\Pi_t,\fmagd{f_t}_{\mathcal K_t}\leq1}{\frac1n\sum_{i=1}^n\fprns{\frac{\delta_{T_it}}{\logging_t(X_i)}-1}\pi_t(X_i)f_t(X_i)}
+\Gamma\sup\limits_{\pi\in\Pi}\abs{1-Z(\pi)^{-1}}.
\end{talign*}
We first treat the random variable
$$\ts
\Xi_t(X_{1:n},T_{1:n})=\sup_{\pi_t\in\Pi_t,\fmagd{f_t}_{\mathcal K_t}\leq1}\frac1n\sum_{i=1}^n\fprns{\frac{\delta_{T_it}}{\logging_t(X_i)}-1}\pi_t(X_i)f_t(X_i).
$$
Fix $x_{1:n},t_{1:n},x'_{1:n},t'_{1:n}$ such that $x_i'=x_i,t_i'=t_i\,\forall i\neq i'$ and note that
\begin{talign*}
\Xi_t(x_{1:n},t_{1:n})-\Xi_t(x'_{1:n},t'_{1:n})
\leq
\sup_{\pi_t\in\Pi_t,\fmagd{f_t}_{\mathcal K_t}\leq1}\bigl(&\frac1n\sum_{i=1}^n\fprns{\frac{\delta_{t_it}}{\logging_t(x_i)}-1}\pi_t(x_i)f_t(x_i)\\&-\frac1n\sum_{i=1}^n\fprns{\frac{\delta_{t'_it}}{\logging_t(x'_i)}-1}\pi_t(x'_i)f_t(x'_i)\bigr)
\tendl=
\frac1n\sup_{\pi_t\in\Pi_t,\fmagd{f_t}_{\mathcal K_t}\leq1}
\fprns{
\fprns{\frac{\delta_{t_{i'}t}}{\logging_t(x_{i'})}-1}\pi_t(x_{i'})f_t(x_{i'})
&-
\fprns{\frac{\delta_{t'_{i'}t}}{\logging_t(x'_{i'})}-1}\pi_t(x'_{i'})f_t(x'_{i'})
}
\leq\frac{2}{n}\alpha\Gamma.
\end{talign*}
By McDiarmid's inequality, $\Prb{{\Xi_t(X_{1:n},T_{1:n})}\geq\Efb{\Xi_t(X_{1:n},T_{1:n})}+\eta}\leq e^{-n\eta^2\alpha^{-2}\Gamma^{-2}/2}$.
Let $\xi_i(\pi_t,f_t)=\fprns{\frac{\delta_{T_it}}{\logging_t(X_i)}-1}\pi_t(X_i)f_t(X_i)$ and note that for all $\pi_t,f_t$ we have
$\Efb{\xi_i(\pi_t,f_t)}=\Efb{\fprns{\Efb{\delta_{T_it}/\logging_t(X_i)\mid X_i}-1}\pi_t(X_i)f_t(X_i)}=0$ and that $\xi_1(\cdot,\cdot),\xi_2(\cdot,\cdot),\dots$ are iid. Therefore, letting $\xi_1'(\cdot,\cdot),\xi_2'(\cdot,\cdot),\dots$  be iid replicates of $\xi_1(\cdot,\cdot),\xi_2(\cdot,\cdot),\dots$ (ghost sample) and letting $\rho_i$ be iid Rademacher random variables independent of all else, we have
\begin{talign*}
\Efb{\Xi_t(X_{1:n},T_{1:n})}
&=
\Efb{\sup_{\pi_t\in\Pi_t,\fmagd{f_t}_{\mathcal K_t}\leq1}{\frac1{n}\sum_{i=1}^n(\Efb{\xi'_i(\pi_t,f_t)}-\xi_i(\pi_t,f_t))}}
\tendl&\leq
\Efb{\sup_{\pi_t\in\Pi_t,\fmagd{f_t}_{\mathcal K_t}\leq1}{\frac1{n}\sum_{i=1}^n(\xi'_i(\pi_t,f_t)-\xi_i(\pi_t,f_t))}}
\tendl&=
\Efb{\sup_{\pi_t\in\Pi_t,\fmagd{f_t}_{\mathcal K_t}\leq1}{\frac1{n}\sum_{i=1}^n\rho_i(\xi'_i(\pi_t,f_t)-\xi_i(\pi_t,f_t))}}
\tendl&\leq
2\Efb{\sup_{\pi_t\in\Pi_t,\fmagd{f_t}_{\mathcal K_t}\leq1}{\frac1{n}\sum_{i=1}^n\rho_i\xi_i(\pi_t,f_t)}}.
\end{talign*}
Note that by bounded kernel we have
$\fmagd{\mathcal K_t(x,\cdot)}_{\mathcal K_t}=\sqrt{\mathcal K_t(x,x)}\leq\Gamma$
and therefore
$$\ts
\sup_{\fmagd{f_t}_{\mathcal K_t}\leq1,x\in\mathcal X}f_t(x)=
\sup_{\fmagd{f_t}_{\mathcal K_t}\leq1,x\in\mathcal X}\ip{f_t}{\mathcal K_t(x,\cdot)}\leq
\sup_{\fmagd{f_t}_{\mathcal K_t}\leq1,\fmagd{g}_{\mathcal K_t}\leq\Gamma}\ip{f_t}{g}=
\Gamma.
$$
As before, $\fmagd{\xi_1-\xi_2}_{\mathcal K_t}^2+\fmagd{\xi_1+\xi_2}^2_{\mathcal K_t}=2\fmagd{\xi_1}^2_{\mathcal K_t}+2\fmagd{\xi_2}^2_{\mathcal K_t}+2\ip{\xi_1}{\xi_2}-2\ip{\xi_1}{\xi_2}=2\fmagd{\xi_1}^2_{\mathcal K_t}+2\fmagd{\xi_2}^2_{\mathcal K_t}$ implies
by induction that $\sum_{\rho_i\in\{-1,+1\}^n}\fmagd{\sum_{i=1}^n\rho_i\xi_i}_{\mathcal K_t}^2=2^n\sum_{i=1}^n\fmagd{\xi_i}_{\mathcal K_t}^2$. Hence, %
\begin{talign*}
\Efb{\fmagd{\frac1{n}\sum_{i=1}^n\rho_iE_{X_i}}_{_{\mathcal K_t}}}
\leq&
\fprns{\Efb{\fmagd{\frac1{n}\sum_{i=1}^n\rho_iE_{X_i}}^2_{_{\mathcal K_t}}}}^{1/2}
=
\fprns{\frac1{n^2}\sum_{i=1}^n\Efb{\fmagd{E_{X_i}}^2_{_{\mathcal K_t}}}}^{1/2}
\leq \Gamma/\sqrt{n}.
\end{talign*}
Note that $\fabs{\frac{\delta_{T_it}}{\logging_t(X_i)}-1}\leq\alpha$,
that $x^2$ is $2b$-Lipschitz on $[-b,b]$, and that $ab=\frac12((a+b)^2-a^2-b^2)$.
Therefore,
by the Rademacher comparison lemma \citep[Thm.~4.12]{ledoux1991probability}, we have
\begin{talign*}
\Efb{\Xi_t(X_{1:n},T_{1:n})}
\leq&
2\alpha\Efb{\sup_{\pi_t\in\Pi_t,\fmagd{f_t}_{\mathcal K_t}\leq1}{\frac1{n}\sum_{i=1}^n\rho_i\pi_t(X_i)f_t(X_i)}}
\tendl\leq&
\alpha\Efb{\sup_{\pi_t\in\Pi_t,\fmagd{f_t}_{\mathcal K_t}\leq1}{\frac1{n}\sum_{i=1}^n\rho_i(\pi_t(X_i)+f_t(X_i))^2}}
\tendl&+
\alpha\Efb{\sup_{\pi_t\in\Pi_t}{\frac1{n}\sum_{i=1}^n\rho_i\pi_t(X_i)^2}}
+
\alpha\Efb{\sup_{\fmagd{f_t}_{\mathcal K_t}\leq1}{\frac1{n}\sum_{i=1}^n\rho_if_t(X_i)^2}}
\tendl\leq&
4\Gamma\alpha\Efb{\sup_{\pi_t\in\Pi_t,\fmagd{f_t}_{\mathcal K_t}\leq1}{\frac1{n}\sum_{i=1}^n\rho_i(\pi_t(X_i)+f_t(X_i))}}
\tendl&+
2\alpha\Efb{\sup_{\pi_t\in\Pi_t}{\frac1{n}\sum_{i=1}^n\rho_i\pi_t(X_i)}}
+
2\Gamma\alpha\Efb{\sup_{\fmagd{f_t}_{\mathcal K_t}\leq1}{\frac1{n}\sum_{i=1}^n\rho_if_t(X_i)}}
\tendl\leq&
6\Gamma\alpha\fprns{{\mathfrak R}_n(\Pi_t)+\Gamma/\sqrt{n}}.
\end{talign*}
Next, let $\omega_{ti}(\pi_t)=(\delta_{T_it}/\logging_{T_it}-1)\pi_t(X_i)$ and $\Omega_t(X_{1:n},T_{1:n})=\sup_{\pi_t\in\Pi_t}\frac1n\sum_{i=1}^n\omega_{ti}(\pi_t)$. Note that $\sup_{\pi\in\Pi}(Z(\pi)-1)\leq \sum_{t=1}^m\Omega_t(X_{1:n},T_{1:n})$.
Fix $x_{1:n},t_{1:n},x'_{1:n},t'_{1:n}$ such that $x_i'=x_i,t_i'=t_i\,\forall i\neq i'$ and note that
\begin{talign*}
\Omega_t(x_{1:n},t_{1:n})-\Omega_t(x'_{1:n},t'_{1:n})
\leq
\frac1n\sup_{\pi_t\in\Pi_t}
\fprns{
\fprns{\frac{\delta_{t_{i'}t}}{\logging_t(x_{i'})}-1}\pi_t(x_{i'})
&-
\fprns{\frac{\delta_{t'_{i'}t}}{\logging_t(x'_{i'})}-1}\pi_t(x'_{i'})
}\leq \frac2n\alpha
\end{talign*}
By McDiarmid's inequality, $\Prb{{\Omega_t(X_{1:n},T_{1:n})}\geq\Efb{\Omega_t(X_{1:n},T_{1:n})}+\eta}\leq e^{-n\eta^2\alpha^{-2}/2}$.
Note that
$\Efb{\omega_{ti}(\pi_t)}=0$ for all $\pi_t$ and that $\omega_{t1}(\cdot),\omega_{t2}(\cdot),\dots$ are iid.
Using the same argument as before, letting $\rho_i$ be iid Rademacher random variables independent of all else, we have
$$\ts\Efb{\Omega_t(X_{1:n},T_{1:n})}
\leq2\Efb{\sup_{\pi_t\in\Pi_t}\frac1n\sum_{i=1}^n\rho_i\omega_{ti}(\pi_t)}
\leq2\alpha{\mathfrak R}_n(\Pi_t).
$$
With a symmetric argument, 
letting $\delta=3m\nu/(3m+2)$,
with probability at least $1-2\delta/3$, we have $\sup_{\pi\in\Pi}\abs{1-Z(\pi)}\leq 2\alpha{\mathfrak R}_n(\Pi)+\alpha\sqrt{{2\log(3m/\delta)}/n}\leq
2\alpha{\mathfrak R}_n(\Pi)+\alpha\sqrt{{2\log(4m/\nu)}/n}\leq1/2$. 

Since $\fmagd{\tilde W}_2\leq \sqrt{n}\alpha/Z(\pi)$, we get that, with probability at least $1-\delta$, both $\sup_{\pi\in\Pi}\fmagd{\tilde W}_2\leq 2\alpha\sqrt{n}$ and for all $t\in[m]$
\begin{talign*}
\sup_{\pi\in\Pi}\mathfrak B_t(\tilde W,\pi_t;\fmagd\cdot_{\mathcal K_t})
\leq&\alpha\Gamma\fprns{12{\mathfrak R}_n(\Pi_t)+2{\mathfrak R}_n(\Pi)+12\Gamma/\sqrt{n}+3\sqrt{2\log(3m/\delta)/n}}.
\end{talign*}
Therefore, with probability at least $1-\delta$, using twice that $\ell_1$ is the biggest $p$-norm,
\begin{talign*}
\mathcal E=\sup_{\pi\in\Pi}\mathfrak E&(W_n^*,\pi;\fmagd \cdot_{p,\mathcal K_{1:m},\gamma_{n,1:m}},\Lambda_n)\leq
\sup_{\pi\in\Pi}\mathfrak E(\tilde W,\pi;\magd \cdot_{p,\mathcal K_{1:m},\gamma_{n,1:m}},\Lambda_n)
\tendl&\leq
\sum_{t=1}^m\gamma_t\sup_{\pi\in\Pi}\mathfrak B_t(\tilde W,\pi_t;\fmagd\cdot_{\mathcal K_t})+
\frac{\overline\kappa}{n}\sup_{\pi\in\Pi}\fmagd{\tilde W}_2
\tendl&\leq
8\alpha\Gamma\overline\gamma m{\mathfrak R}_n(\Pi)
+\frac{2\alpha\overline\kappa+{12\alpha\Gamma^2\overline\gamma m+3\alpha\Gamma\overline\gamma m\sqrt{2\log(3m/\delta)}}}{\sqrt{n}}.
\end{talign*}
Consider case \ref{uniformconsistencythm:finitenorm}. Note that 
$\sup_{\pi\in\Pi}\sum_{t=1}^m\fabs{B_t(W_n^*,\pi_t;\mu_t)}\leq\magd{\mu}\mathcal E$ and
$\sup_{\pi\in\Pi}\fmagd{W_n^*}_2\leq \underline\kappa^{-1}\mathcal E$. Since $\Efb{\sum_{i=1}^nW_i\epsilon_i\mid \XTs}=0$, $\epsilon_i\in[-B,B]$ 
and
$W_i\epsilon'_i-W_i\epsilon''_i\leq 2BW_i$ for $\epsilon'_i,\epsilon''_i\in[-B,B]$, by
McDiarmid's inequality (conditional on $\XTs$), we have that 
with probability at least $1-\delta'$,
$\fabs{\sum_{i=1}^nW_{ni}^*\epsilon_i}\leq
\fmagd {W_n^*}_2B\sqrt{2\log(2/\delta)}
.$
Therefore, letting 
$\delta'=2\nu/(3m+2)$ so that $3m/\delta=2/\delta'=(3m+2)/\nu\leq4m/\nu$, 
with probability at least $1-\nu$, we have
\begin{talign*}
&\sup_{\pi\in\Pi}\fabs{\tau_{W_n^*}-\op{SAPE}(\pi)}\leq
8\alpha\Gamma\overline\gamma m(\fmagd\mu+\sqrt{2\log(4m/\nu)}\underline\kappa^{-1}B){\mathfrak R}_n(\Pi)
\tendl&+\frac{
{2\alpha\overline\kappa\fmagd\mu+12\alpha\Gamma^2\overline\gamma m\fmagd\mu}
+
\fprns{
2\alpha\overline\kappa\underline\kappa^{-1}B+
12\alpha\Gamma^2\overline\gamma m\underline\kappa^{-1}B+
3\alpha\Gamma\overline\gamma m\fmagd\mu}\sqrt{2\log(4m/\nu)}
+
6\alpha\Gamma\overline\gamma m\underline\kappa^{-1}B{\log(4m/\nu)}
}
{\sqrt{n}}.
\end{talign*}
This gives the first result in case \ref{uniformconsistencythm:finitenorm}. The second is given by noting that, by McDiarmid's inequality, with probability at least $1-\nu/(4m)$,
${\mathfrak R}_n(\Pi_t)\leq\widehat{\mathfrak R}_n(\Pi_t)+4\sqrt{2\log(4m/\nu)}$.
Case \ref{uniformconsistencythm:univ} is given by following a similar argument as in the proof
of Thm.~\ref{consistencythm}\ref{consistencythm:univ}.
\end{proof}

\begin{proof}[Proof of Cor.~\ref{uniformconsistencycor1}]
These results follow directly from the proof of Thm.~\ref{uniformconsistencythm},
the convergence in particular of $\mathfrak E^2(W^*(\pi),\pi;\fmagd\cdot,\Lambda)$,
the decomposition of the DR estimator in Thm.~\ref{cmsethm},
and a standard Rademacher complexity argument concentrating $\op{SAPE}(\pi)$ 
uniformly around $\op{PAPE}(\pi)$.
\end{proof}

\section{IPW and DR weight SVM details}

To reduce training a deterministic linear policy using IPW evaluation to weighted
SVM classification, we add multiples of $\sum_{i=1}^n\pi_{T_i}(X_i)/\hat\phi_{T_i}(X_i)$
(1 in expectation) and note that
\begin{talign*}
\frac1B\fprns{\hat\tau^\text{IPW}(\pi)-C\sum_{i=1}^n\frac{\pi_{T_i}(X_i)}{\hat\phi_{T_i}(X_i)}-\sum_{i=1}^n\frac{Y_i-C}{\hat\phi_{T_i}(X_i)}}
&=\sum_{i=1}^n\frac{C-Y_i}{B\hat\phi_{T_i}(X_i)}(1-\pi_{T_i}(X_i))
\tendl&=\sum_{i=1}^n\frac{C-Y_i}{B\hat\phi_{T_i}(X_i)}\findic{T_i\neq\tilde T_{\pi(X_i)}}.
\end{talign*}
Choosing $C$ sufficiently large so that all coefficients are nonnegative and choosing
$B$ so that all coefficients are in $[0,1]$, we replace the indicators 
$\findic{T_i\neq\tilde T_{\pi(X_i)}}$ with 
their convex envelope hinges to come up with
a weighted version of \citet{crammer2001algorithmic}'s multiclass SVM.

For the DR version, we replace $\pi_t(X_i)$ with $\findic{t=\tilde T_{\pi(X_i)}}$
and we do the above with but using $\hat\epsilon_i$
and also add multiples of 
$\hat\tau^\text{direct}(1(\cdot))=\sum_{i=1}^n\sum_{t=1}^m\mu_t(X_i)$
to make all indicators be 0-1 loss and have nonnegative coefficients.
Replacing indicators with hinge functions, we get a weighted multiclass SVM
with different weights for each observation and \emph{each error} type.

\end{document}